\documentclass[11pt]{article}

\usepackage[preprint]{acl}

\usepackage{times}
\usepackage{latexsym}
\usepackage[T1]{fontenc}
\usepackage[utf8]{inputenc}
\usepackage{microtype}
\usepackage{inconsolata}

\usepackage{graphicx}
\usepackage{booktabs}
\usepackage{amsfonts}
\usepackage{amsmath}
\usepackage{multirow}
\usepackage{float}
\usepackage{xcolor}
\usepackage{subcaption}
\usepackage{enumitem}

\newcommand{\Paragraph}[1]{\noindent\textbf{#1}}

\title{HydroAgent: Closing the Gap Between Frontier LLMs and Human Experts in Hydrologic Model Calibration via Simulator-Grounded RL}

\author{
  \textbf{Zhi Li\textsuperscript{1}},
  \textbf{Songkun Yan\textsuperscript{2}},
  \textbf{Jie Cao\textsuperscript{3}},
  \textbf{Mofan Zhang\textsuperscript{4}},
  \textbf{Anjiang Wei\textsuperscript{5}},
  \textbf{Jinwoong Yoo\textsuperscript{6}},
  \textbf{Yang Hong\textsuperscript{2}}
  \\
  \textsuperscript{1}Civil, Environmental, and Architectural Engineering, University of Colorado Boulder \\
  \textsuperscript{2}Civil Engineering and Environmental Sciences, University of Oklahoma \\
  \textsuperscript{3}Department of Computer Science, University of Oklahoma \\
  \textsuperscript{4}Civil and Environmental Engineering, Stanford University \\
  \textsuperscript{5}Department of Computer Science, Stanford University \\
  \textsuperscript{6}NASA Goddard Space Flight Center
  \\
  \small{
    \textbf{Correspondence:} \href{mailto:Zhi.Li-2@colorado.edu}{Zhi.Li-2@colorado.edu}
  }
}

\begin{document}
\maketitle

\begin{abstract}
Calibrating distributed hydrologic models is a critical bottleneck across operational water-resources management---streamflow prediction, water-supply assessment, reservoir operation, drought monitoring, infrastructure design, and flood forecasting all depend on it: each basin demands a domain expert to translate hydrograph signatures into adjustments of a high-dimensional parameter vector, and the resulting workflow does not transfer between watersheds. We ask a sharper version of a now-common question: \emph{can frontier large language model (LLM) agents replace the human hydrologic modeler, and if not, what would it take?} We benchmark nine frontier LLM agents---Claude Opus 4.6/4.7, Sonnet 4.6, GPT-5/5.4/5.4-pro, and Gemini 2.5-pro/3.1-pro/3-flash---on the calibration of the operational CREST distributed hydrologic model used by the U.S.\ National Weather Service for flash-flood forecasting. Mean best-of-twenty-rounds Nash--Sutcliffe Efficiency (NSE) across four held-out gauges spanning $329$--$40{,}792$\,km$^2$ ranges from $-0.16$ (GPT-5.4) to $0.75$ (Sonnet 4.6); the ceiling is reproducible across all three frontier vendors and across capability tiers --- the strongest models concentrate in the $0.65$--$0.75$ band, and no model reaches the human-expert reference except for Opus-4.7 at one testing gauge. We argue this gap is not a parameter-count problem: it is a domain-grounding problem. We then propose \textsc{HydroAgent}, a recipe that fine-tunes the open-weight Qwen3-4B model with supervised fine-tuning on $2{,}576$ expert calibration trajectories and Group-Relative Policy Optimization using NSE as a verifiable reward sourced from online CREST simulations---reinforcement learning with simulation feedback (RLSF). Our central thesis is that, for Earth-system science, a small domain-tuned policy with simulator-in-the-loop RL is a more compute-efficient and physically faithful path than scaling generic frontier models, and that the multi-modal richness of Earth data---remote sensing, in-situ time series, and forecaster narrative---makes domain agents an unusually leveraged direction for AI in physical science.
\end{abstract}

\section{Introduction}
\label{sec:intro}

Hydrologic models are the infrastructure layer beneath a remarkable amount of public life---they are how human society manages water. Every reservoir release a water utility schedules, every irrigation allotment a basin authority issues, every drought outlook a farmer plans against, every hydropower bid a grid operator submits, every wastewater-treatment design a city engineer approves, and every flash-flood watch a forecaster issues derives from a numerical rainfall-runoff model that has been calibrated, in advance, against historical streamflow at the relevant gauge. The stakes are not abstract: flooding is the deadliest weather hazard in the United States \citep{li2021flood} and among the costliest natural disasters globally, with more than $1.8$ billion people directly exposed to one-in-hundred-year flood risk and the burden falling disproportionately on the poorest populations \citep{rentschler2022flood,mcdermott2022global}, and accurate, timely streamflow forecasts are the difference between an evacuation that works and one that does not. The same models also price national flood insurance, set the design standards of bridges, levees, and stormwater systems, and tell municipalities how much water they can deliver through the dry months. As climate change accelerates the hydrologic cycle and pushes flash floods and droughts into regions that have never experienced them at current intensities \citep{li2022conterminous}, demand for accurate, basin-specific simulations is rising at exactly the moment when the supply of trained hydrologic modelers cannot keep pace. AI is the obvious candidate to absorb that workload; the question this paper asks is \emph{how}.

Hydrologic model calibration is the act of choosing a small number of physical parameters---soil-water capacity, infiltration shape, channel-routing coefficients---so that a numerical rainfall-runoff model reproduces observed streamflow at a stream gauge. It is the human-in-the-loop step that makes every downstream use of the model trustworthy---reservoir operation, water-supply allocation, drought outlook, hydropower scheduling, infrastructure design, and flood warning alike---and it is the step that does \emph{not} scale: each gauge in the conterminous United States requires hours to days of an expert hydrologist's attention, and what they learn about one basin transfers only loosely to the next. The mismatch between the climate-driven rise in demand sketched above and the human supply of calibration expertise is the gap that motivates this paper.

The recent surge in agentic large language models---models that interleave chain-of-thought reasoning with tool use over many turns \citep{yao2023react,schick2023toolformer,qin2024toolllm}---raises an obvious question. These models have absorbed graduate-level hydrology textbooks, can read CSV gauge files, can edit control files, and can shell out to a Linux binary. Can they simply \emph{do} the calibration? If yes, the cost structure of operational hydrology shifts overnight. If no, the manner of failure tells us where the gap actually lives.

This paper investigates that question along two complementary axes.

\Paragraph{Axis 1: Where do frontier agents stand?} We give nine frontier LLM agents---Claude Opus 4.6, Opus 4.7, Sonnet 4.6, GPT-5, GPT-5.4, GPT-5.4-pro, Gemini 2.5-pro, Gemini 3.1-pro, and Gemini 3-flash---the same task an entry-level hydrologist receives: a basin description, gauge time series, raster forcings (MRMS gauge-corrected precipitation, daily PET), the CREST control-file template, and the goal of achieving high NSE values. Each agent runs the calibration independently on each of four held-out gauges spanning $329$--$40{,}792$\,km$^2$, inside the same Linux sandbox under the \emph{terminal-2} agent harness in the \emph{harbor} evaluation framework with a fixed compute budget (2-hour timeout); the same harness is also used to evaluate our fine-tuned model in Section~\ref{sec:hydrollm}, so frontier-model and \textsc{HydroAgent} numbers are directly comparable.

\begin{figure*}[t]
  \centering
  \includegraphics[width=0.95\linewidth]{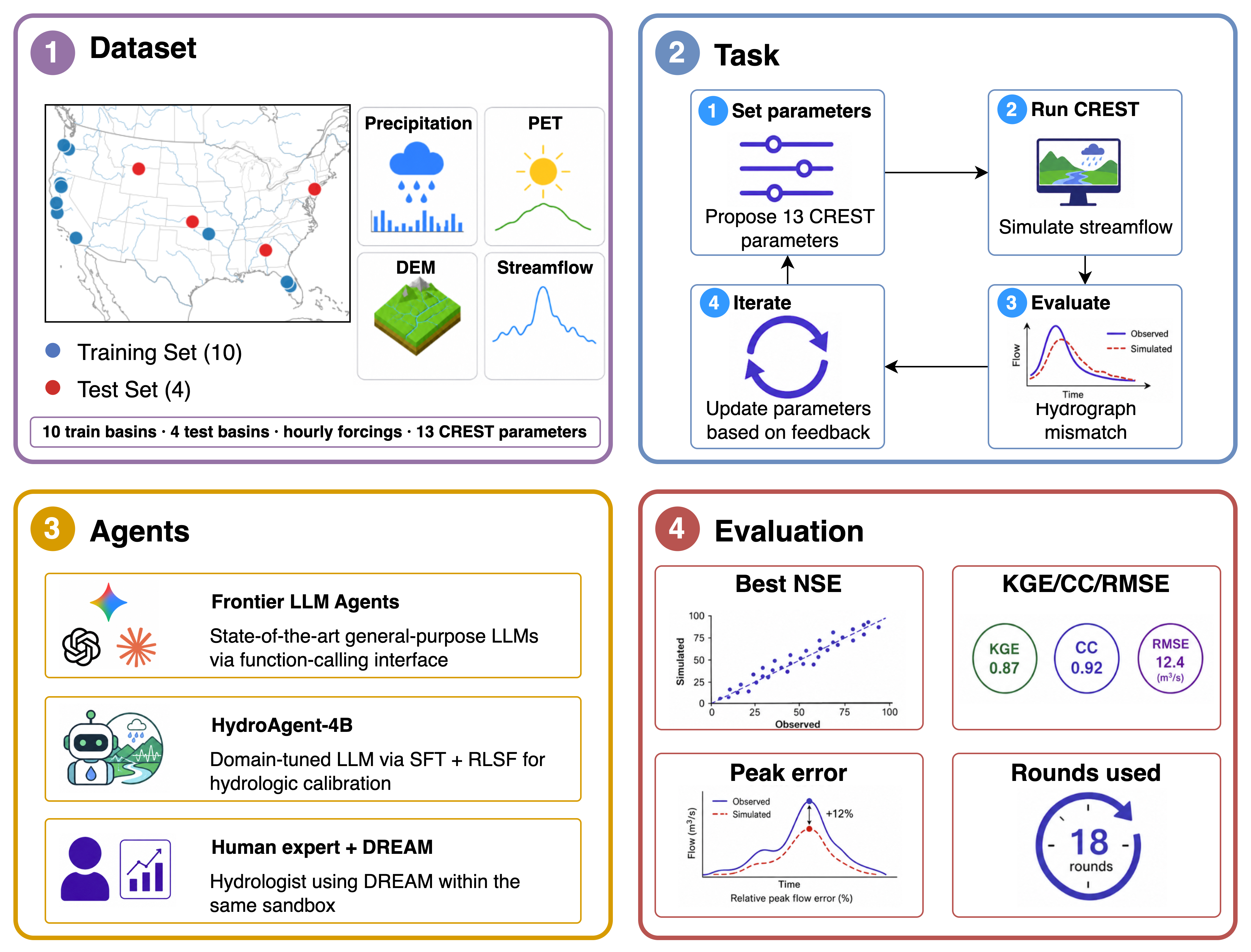}
  \caption{Overview of the hydrologic calibration benchmark and evaluation pipeline: Dataset, Task, Agents, Evaluations. Phase~1: supervised fine-tuning (SFT) on $2{,}576$ expert calibration trajectories distills tool-call format and basic hydrologic reasoning into Qwen3-4B. Phase~2: Group-Relative Policy Optimization (GRPO) draws $K{=}8$ rollouts per prompt; each rollout proposes a CREST parameter set, executes the EF5 hydrologic simulator, scores a clipped Nash--Sutcliffe Efficiency, and updates the policy with group-normalized advantages. The simulator is the verifier: there is no learned reward model.}
  \label{fig:schematic}
\end{figure*}

\Paragraph{Axis 2: Is a small, simulator-grounded model enough?} Frontier-model evaluation suggests that the challenge is not raw reasoning capacity but \emph{calibration grounding}: the model has not learned which parameter to move when the recession limb is too steep. To test whether that gap can be narrowed without scaling, we propose \textsc{HydroAgent}, a domain-specific agent built on the open-weight Qwen3-4B-Instruct model \citep{qwen2025qwen3}. As Figure~\ref{fig:schematic}, training proceeds in two phases: (i) supervised fine-tuning (SFT) on $2{,}576$ calibration trajectories distilled from a stronger teacher, and (ii) reinforcement learning~(RL) with simulation feedback~(RLSF) using Group-Relative Policy Optimization~(GRPO) \citep{shao2024deepseekmath,deepseek2025r1}, where each rollout invokes an online CREST simulation and the reward is a clipped NSE plus shaped per-turn signals.

The argument we want to defend in this paper has three pieces. First, frontier LLM agents are within striking distance of human-expert performance on hydrologic-model calibration but do not yet meet it; the gap is reproducible across model families. Second, that gap closes faster by \emph{post-training a small open model with simulator-in-the-loop RL} than by waiting for the next frontier release; one does not need a 400-billion-parameter generalist to operate a 13-parameter physics simulator, and recent work on agentic AI for Earth observation has independently argued that generic agent recipes leave structural problems---geospatial consistency, physical validity, error propagation across long tool chains---unaddressed in ways that domain-tuned designs can fix \citep{munir2026agenticrs}. Third, Earth-system science is an unusually fertile ground for this recipe because the data are rich and multi-modal: rasterized remote-sensing imagery (spatial-2D + time), in-situ measurements (time series), and expert narrative (e.g.,~National Weather Service warning messages and forecaster discussions) all carry information that a domain agent can be trained to fuse with a physical solver to yield steerable, physically consistent predictions.

\Paragraph{Contributions.} (1) The first systematic benchmark of nine frontier LLM agents on the calibration of an operational distributed hydrologic model, including a public release of all $9 \times 20$ trajectories for reproducibility (Section~\ref{sec:frontier}). (2) A domain-specific recipe (\textsc{HydroAgent}) that fine-tunes Qwen3-4B with SFT + GRPO using NSE as a verifiable simulator-grounded reward, with full training configuration sufficient to reproduce the run on $4\times$H100 (Section~\ref{sec:methods}). (3) A position---defended quantitatively---that for Earth-system tasks with cheap-to-evaluate physical simulators, a domain-tuned 4B-parameter agent can substantially close the gap to frontier generalists (Section~\ref{sec:hydrollm}).

\section{Methods}
\label{sec:methods}

This section describes (i) the CREST hydrologic simulator that defines our task and reward, (ii) the calibration environment exposed to the agent, and (iii) the SFT~+~RLSF training recipe used to produce \textsc{HydroAgent}. The training infrastructure is \texttt{verl}~0.5 \citep{sheng2024hybridflow} with SGLang multi-turn rollouts on $4\times$ NVIDIA H100 (80\,GB) GPUs; configurations sufficient to reproduce the run are given in Appendix~\ref{app:repro}.

\subsection{The CREST/EF5 Hydrologic Simulator}
\label{sec:crest}

The Coupled Routing and Excess STorage (CREST) model \citep{wang2011coupled,li2023crestreview} is a distributed, grid-based, conceptual rainfall-runoff model. At each grid cell and each hourly time step, CREST partitions incoming precipitation into impervious runoff, infiltration, and evapotranspiration as a function of a soil-water capacity grid (\texttt{wm}), a variable infiltration-curve shape parameter (\texttt{b}), an impervious-area fraction (\texttt{im}), an evapotranspiration scaling (\texttt{ke}), and a saturated hydraulic conductivity (\texttt{fc}); routes interflow with leakage (\texttt{leaki}) and velocity (\texttt{under}); and routes channel and overland flow with kinematic-wave parameters $\alpha$, $\beta$, and $\alpha_0$. Initial soil moisture (\texttt{iwu}) and a small set of state parameters complete the parameter vector; spatial heterogeneity is encoded in raster grids, and per-basin calibration is performed by tuning a vector of \emph{scalar multipliers} on those grids, which preserves the spatial pattern while shifting magnitudes.

CREST is hosted by the Ensemble Framework For Flash Flood Forecasting (EF5) \citep{flamig2020ef5}, which is the operational hydrologic engine of the FLASH project \citep{gourley2017flash} at the NOAA National Severe Storms Laboratory and is used in real time by U.S.\ National Weather Service forecasters to issue flash-flood watches and warnings across the CONUS. CREST/EF5 has been deployed globally, including in data-sparse satellite-forced settings \citep{xue2013statistical}, and projections suggest that the demand for accurate calibration of CREST-class models will grow: under SSP5--8.5, flash-flood frequency over the CONUS is projected to increase \citep{li2022conterminous}. Two facts about CREST shape our experimental design: (i) it is non-differentiable in the parameter vector, ruling out gradient-based calibration and motivating an agent-in-the-loop approach, and (ii) a single basin/event simulation completes in seconds-to-minutes, which makes simulator-feedback RL practical at the rollout scales we report.

\subsection{Calibration Environment and Task}
\label{sec:env}

We instantiate one calibration episode per basin/event pair. Each episode exposes four tools to agents:
\begin{itemize}[
    leftmargin=1.2em,
    itemsep=1pt,
    topsep=2pt,
    parsep=0pt,
    partopsep=0pt
]
  \item \texttt{set\_parameters}: write the 13 scalar multipliers (\texttt{wm}, \texttt{b}, \texttt{im}, \texttt{ke}, \texttt{fc}, \texttt{under}, \texttt{leaki}, \texttt{alpha}, \texttt{beta}, \texttt{alpha0}, \texttt{iwu}, \texttt{th}, \texttt{isu}) into a control-file copy. Each parameter has a documented physically-motivated range (e.g., $\texttt{wm}\!\in\![0.1,\,10.0]$, $\texttt{im}\!\in\![0,1]$); the full list of bounds and physical interpretations is given in Appendix~\ref{app:params}.
  \item \texttt{run\_simulation}: invoke the EF5 binary on the patched control file and parse the resulting time-series CSV, returning the full simulated/observed discharge series at hourly resolution along with a summary of multiple hydrologic-fit signatures: peak-flow magnitude error and timing offset, total-volume ratio~(a water-balance closure indicator), recession-limb slope, time-to-peak, baseflow level, and event count. These criteria are exposed as text the LLM interprets jointly---a peak under-prediction with a balanced volume ratio implies a routing-velocity issue, while a balanced peak with a volume surplus implies an evapotranspiration or impervious-fraction issue.
  \item \texttt{evaluate}: aggregate the latest simulation against the gauge time series. Returns the Nash--Sutcliffe Efficiency \citep{nash1970river}, $\mathrm{NSE}=1-\sum_t(Q_t^{\mathrm{obs}}-Q_t^{\mathrm{sim}})^2/\sum_t(Q_t^{\mathrm{obs}}-\overline{Q^{\mathrm{obs}}})^2$, alongside a panel of complementary metrics---Kling--Gupta Efficiency components (correlation, variability ratio, bias ratio), root-mean-square error, percent bias, and high-/low-flow KGE---together with the running best-NSE and target-status. The agent therefore reasons over a \emph{multi-criteria} diagnostic rather than a single scalar; NSE is the headline reward signal but the auxiliary metrics are what disambiguate which physical process is the next thing to adjust.
  \item \texttt{parse\_failure}: understands the difference between model simulated and observed streamflow from its metrics: CC, KGE, NSE, peak\_error, time\_lag. Reasons the root cause and provides possible reasons. An example of the reasoning output is \textit{``The model has stabilized with an NSE of 0.2248, CC of 0.591, and KGE of 0.5588, indicating a well-calibrated simulation that accurately captures the observed hydrograph in terms of timing and volume. Although the peak discharge (374.93) is still below the observed value (622.97), the timing error (lag = 32 hours) is minimal, and the model's performance across all metrics is consistent and physically realistic. Further improvements would require adjusting channel routing or impervious fraction.''}
\end{itemize}

The agent receives the basin descriptor, the gauge directory, the precipitation/PET raster directories, the control-file template, and the evaluation window, and must improve NSE over up to 50 multi-turn iterations. Episodes \textbf{terminate} on target attainment, on five rounds without improvement, or on the per-episode wall-clock budget. We use 10 CONUS gauges spanning $539$--$2401$\,km$^2$ (selected from the audit pool described in Appendix~\ref{app:repro}) for training, and four held-out gauges~(ranging from 329 to 40,792 km$^2$) for evaluation, matching the protocol used in our frontier-model benchmark (Section~\ref{sec:frontier}). We adopt the widely used hydrologic performance rubric of \citet{moriasi2015performance}, which classifies streamflow simulations as \emph{unsatisfactory} ($\mathrm{NSE}\!\leq\!0.50$), \emph{satisfactory} ($0.50\!<\!\mathrm{NSE}\!\leq\!0.70$), \emph{good} ($0.70\!<\!\mathrm{NSE}\!\leq\!0.85$), and \emph{very good} ($\mathrm{NSE}\!>\!0.85$); the $0.8075$ target therefore sits at the top of the \emph{good} band and is the threshold an experienced operational hydrologist routinely meets on this gauge.

\subsection{Stage~1: Supervised Fine-Tuning}
\label{sec:sft}

The base model is \texttt{Qwen3-4B-Instruct-2507} \citep{qwen2025qwen3}, chosen because it is the largest open-weight model that fits comfortably in BF16 with full FSDP fine-tuning and $K{=}8$ multi-turn rollouts on $4\times$H100 (80\,GB), and because it ships with native Hermes-style tool calling. We construct $2{,}576$ SFT trajectories by distilling a strong proprietary teacher (GPT-5\footnote{Due to budget constraints, we use GPT-5 for initial agent development and trajectory collection, while evaluating the four test gauges using nine latest frontier models.}) on $73$ calibration runs across $29$ U.S.\ gauges~(excluding the 4 gauges in the test set). The individual iterative episodes within each calibration run are reorganized into a single long-horizon trajectory --- preserving the full hypothesise $\rightarrow$ simulate $\rightarrow$ diagnose $\rightarrow$ adjust sequence in its original tool-call order --- so that the model learns the chain-of-thought of iterative parameter refinement rather than only the converged parameter set. Trajectories whose final NSE does not exceed $0.6$ are dropped, and the surviving sequences are quality-weighted by their final-NSE percentile. SFT teaches the model the tool-call grammar, the parameter-bounds convention, and a basic hydrologic-reasoning style (``recession limb is too steep $\Rightarrow$ increase \texttt{wm} or \texttt{leaki}'').

\subsection{Stage~2: Reinforcement Learning with Simulation Feedback (RLSF)}
\label{sec:rlsf}

We refer to our RL stage as \emph{Reinforcement Learning with Simulation Feedback} (RLSF) to distinguish it from human-feedback RL \citep{lee2024rlaif} and from pure verifiable-text-reward RL \citep{lambert2024tulu3,deepseek2025r1}: the verifier here is a numerical Earth-system simulator producing continuous physical-error metrics. The optimizer is GRPO \citep{shao2024deepseekmath}, which is critic-free and uses group-relative advantage normalization---a property that fits cheap-but-noisy simulator rewards well. For each prompt we draw $K{=}8$ rollouts; each rollout is a multi-turn calibration episode invoking up to $50$ tool calls. The reward decomposes into per-turn shaping and a terminal score:
\begin{align*}
  r^{\mathrm{turn}}_t &=
  \begin{cases}
    +0.02 & \text{valid \texttt{set\_parameters}} \\
    +0.05 & \text{valid \texttt{run\_simulation}} \\
    \Delta\mathrm{NSE}_t & \text{valid \texttt{evaluate}} \\
    -0.5 & \text{\texttt{parse\_failure}}
  \end{cases}
  \\[2pt]
  r^{\mathrm{terminal}} &= \mathrm{clip}(\mathrm{NSE}^\star,-1,1) \\
  &\quad{}+ 0.5\cdot\mathbf{1}\{\mathrm{NSE}^\star{>}\tau\} + 0.02\,n_{\mathrm{eval}} \\
  &\quad{}+ 0.10\max(0,\,n_{\mathrm{improve}}{-}1) \\
  &\quad{}- 1.0\cdot\mathbf{1}\{\text{empty}\}
\end{align*}
where $\mathbf{1}\{\cdot\}$ is the indicator function (taking value $1$ when the bracketed condition holds and $0$ otherwise), $\Delta\mathrm{NSE}_t$ is the change in best-NSE on turn $t$, $\mathrm{NSE}^\star$ is the episode-best NSE, $\tau$ is the gauge-specific target, $n_{\mathrm{eval}}$ counts valid \texttt{evaluate} calls, $n_{\mathrm{improve}}$ counts evaluations that beat the running best, and an episode is \emph{empty} when the agent never produces a parseable evaluation. The improvement bonus is the key inductive bias: it explicitly trains \emph{iterate-until-you-cannot-improve}, which is what a hydrologist actually does.

We use full BF16 fine-tuning under FSDP (no LoRA), with actor learning rate $1{\times}10^{-6}$, KL anchor coefficient $0.2$ to the SFT initialization, entropy coefficient $0.01$, sampling temperature $1.0$ with top-$p$ $0.95$, $\mathrm{batch}=4$ prompts $\times K{=}8$ rollouts per step, and $30$ epochs over the $10$-gauge training set. The strong KL anchor is necessary: lower values let the policy drift to token-level degenerate outputs by step ${\sim}40$. Rollouts are served by SGLang in synchronous mode with native multi-turn tool dispatch; EF5 invocations are gated by a $32$-way semaphore to prevent CPU/IO contention. Training takes about $5$ hours per checkpoint cadence on $4\times$H100. Full Hydra configurations are in Appendix~\ref{app:repro}.

\section{AI Agents for Hydrologic Modeling: Where Do We Stand?}
\label{sec:frontier}

Hydrologic calibration is a stringent test of agentic scientific reasoning: the task is long-horizon, simulator-grounded, multi-objective, and physically constrained. Effective calibration requires sustained iterative refinement over delayed feedback rather than one-shot reasoning. We therefore evaluate nine frontier LLM agents on a geographically held-out EF5 calibration benchmark~(\S\ref{ssec:benchmark}) to measure both final performance~(\S\ref{ssec:frontier-results}) and long-horizon engagement~(\S\ref{ssec:dynamics}).

\subsection{Benchmark Setup}
\label{ssec:benchmark}

We evaluate nine frontier LLM agents on the calibration of the four held-out gauges of Table~\ref{tab:gauge-config} (drainage areas $329$--$40{,}792$\,km$^2$ across distinct hydroclimatic regions of the CONUS). Each agent runs each gauge~(\S\ref{ssec:frontier-results}) independently inside the same Linux sandbox under the \emph{terminal-2} agent harness in the \emph{harbor} evaluation framework. The harness exposes a single Bash terminal: the agent reads the per-gauge task brief (basin descriptor, parameter table with bounds, EF5 invocation contract, NSE target), inspects the data directory, edits the control file, runs \texttt{ef5 /app/control.txt}, parses the simulated-versus-observed CSV, and iterates. Each round consists of up to $10$ candidate parameter sweeps; the budget is $20$ rounds per gauge (i.e., up to $200$ EF5 simulations per gauge, $800$ across the panel) and we track the running-best NSE per round per gauge. Figure~\ref{fig:bestnse-02338660} visualizes the per-model best-NSE bars on the canonical evaluation gauge \texttt{02338660}. Hydrologic-model calibration is a \emph{long-horizon} task: in our pilot runs, $20$ rounds is the lower end of what suffices to reach the target on a typical gauge, and the multi-criteria diagnostic the agent must integrate (Section~\ref{sec:env}) only becomes informative after several rounds of trial parameter moves. Whether each agent actually \emph{uses} that horizon is itself an outcome of the experiment, reported alongside each best-NSE bar in Figure~\ref{fig:bestnse}.

The nine evaluated models are: \textbf{Anthropic} Claude Opus 4.6, Opus 4.7, and Sonnet 4.6; \textbf{OpenAI} GPT-5, GPT-5.4, and GPT-5.4-pro; \textbf{Google} Gemini 2.5-pro, Gemini 3.1-pro, and Gemini 3-flash. The choice is intended to span the three frontier families and to cover the speed/quality variants currently in production (April~2026). All models are accessed through the harness's default API integration with default sampling temperatures.

\subsection{Results}
\label{ssec:frontier-results}

\begin{figure*}[t]
    \centering
    \begin{subfigure}[t]{0.49\textwidth}
        \centering
        \includegraphics[width=\linewidth]{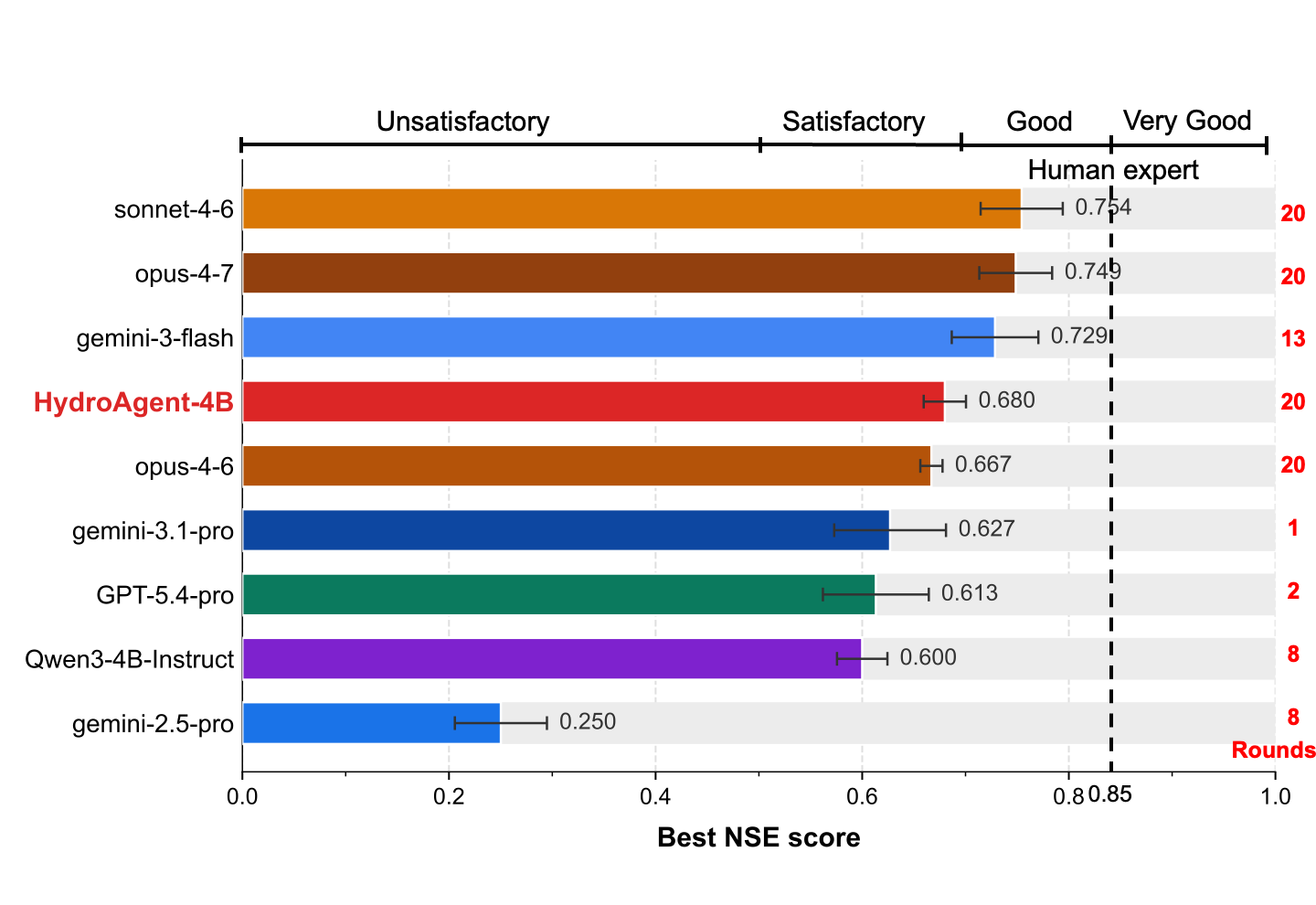}
        \vspace{-2em}
        \caption{\texttt{02338660}}
        \label{fig:bestnse-02338660}
    \end{subfigure}
    \hfill
    \begin{subfigure}[t]{0.49\textwidth}
        \centering
        \includegraphics[width=\linewidth]{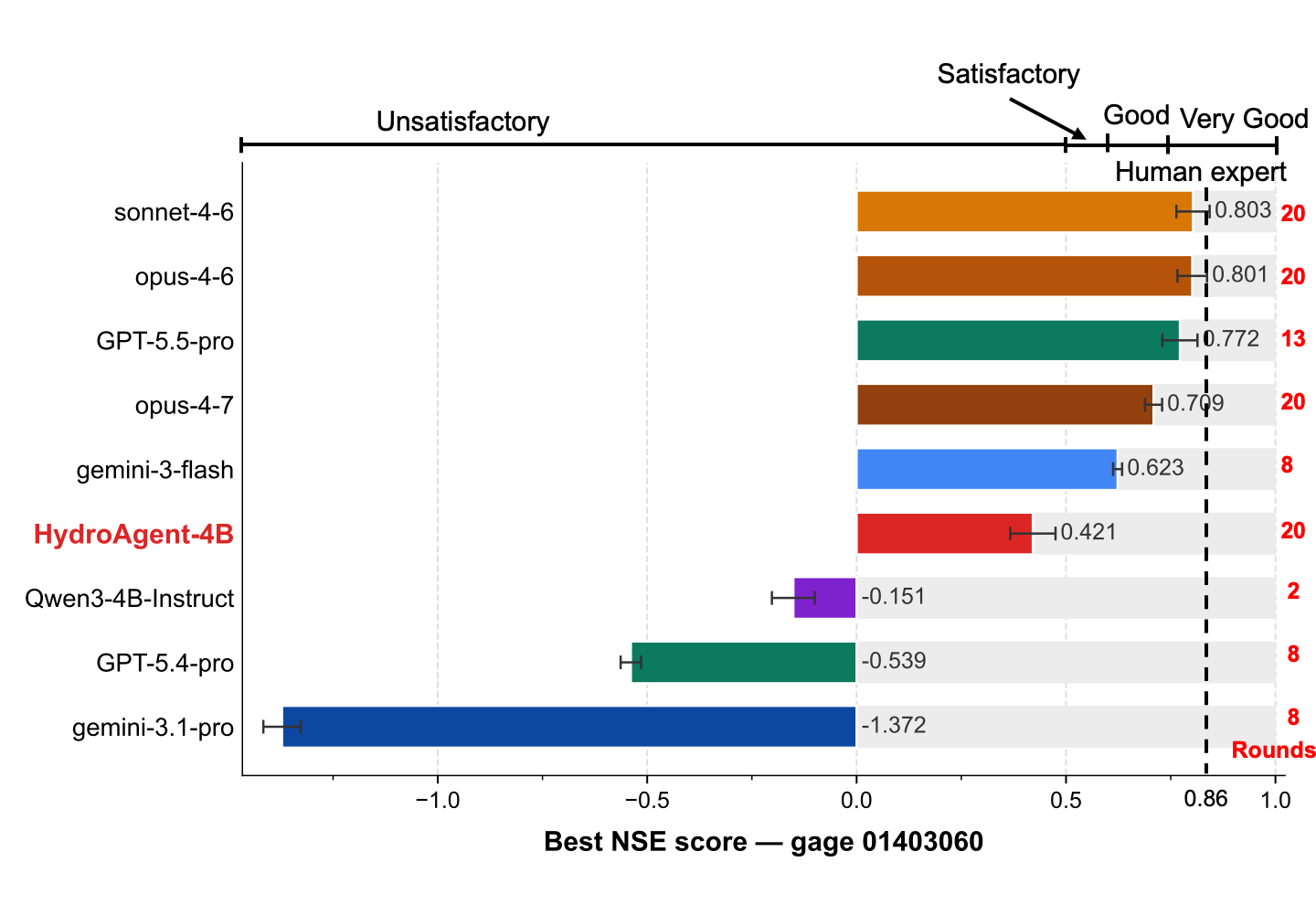}
        \vspace{-2em}
        \caption{\texttt{01403060}}
        \label{fig:bestnse-01403060}
    \end{subfigure}
    \hfill
    \begin{subfigure}[t]{0.49\textwidth}
        \centering
        \includegraphics[width=\linewidth]{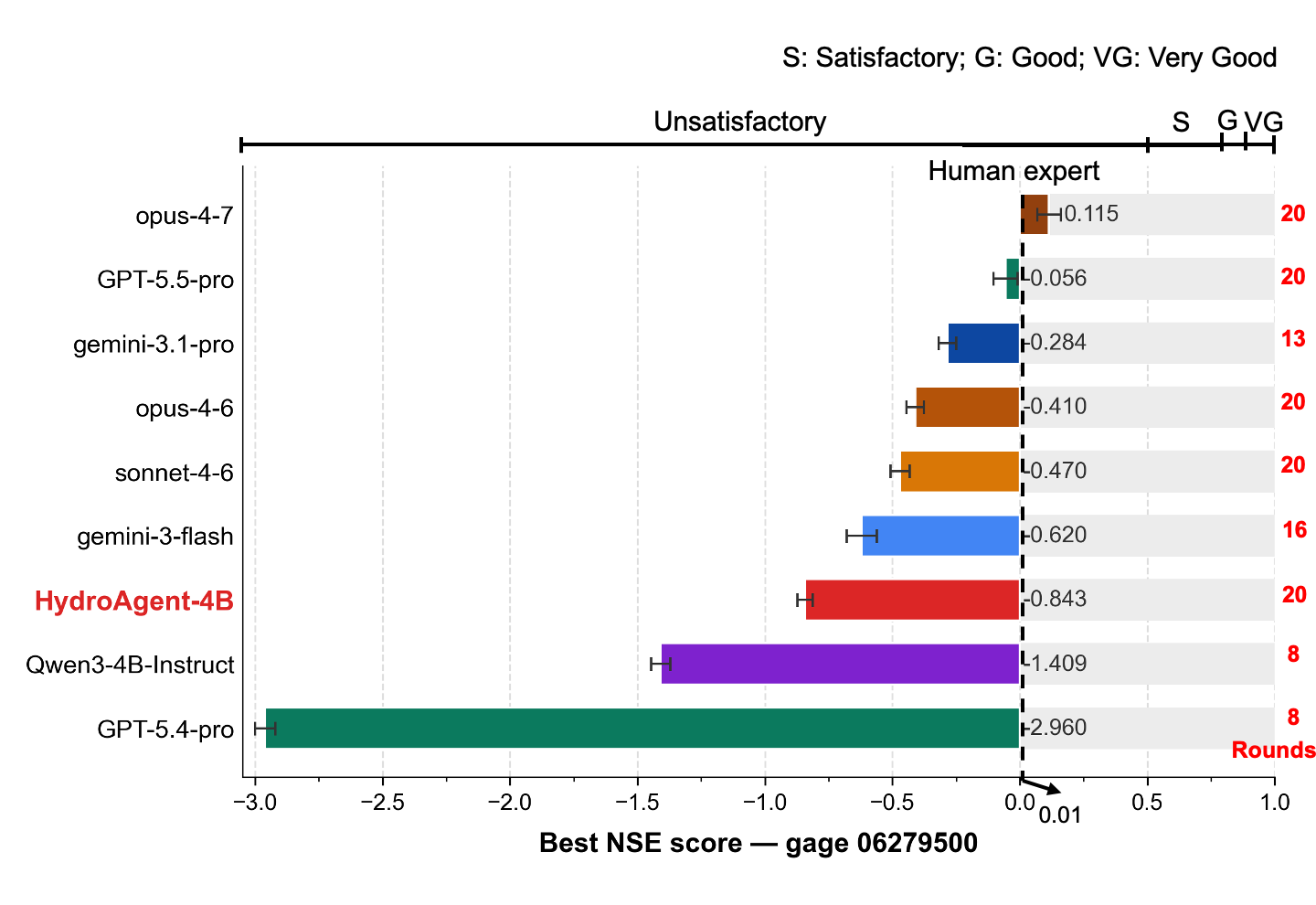}
        \vspace{-2em}
        \caption{\texttt{06279500}}
        \label{fig:bestnse-06279500}
    \end{subfigure}
    \hfill
    \begin{subfigure}[t]{0.49\textwidth}
        \centering
        \includegraphics[width=\linewidth]{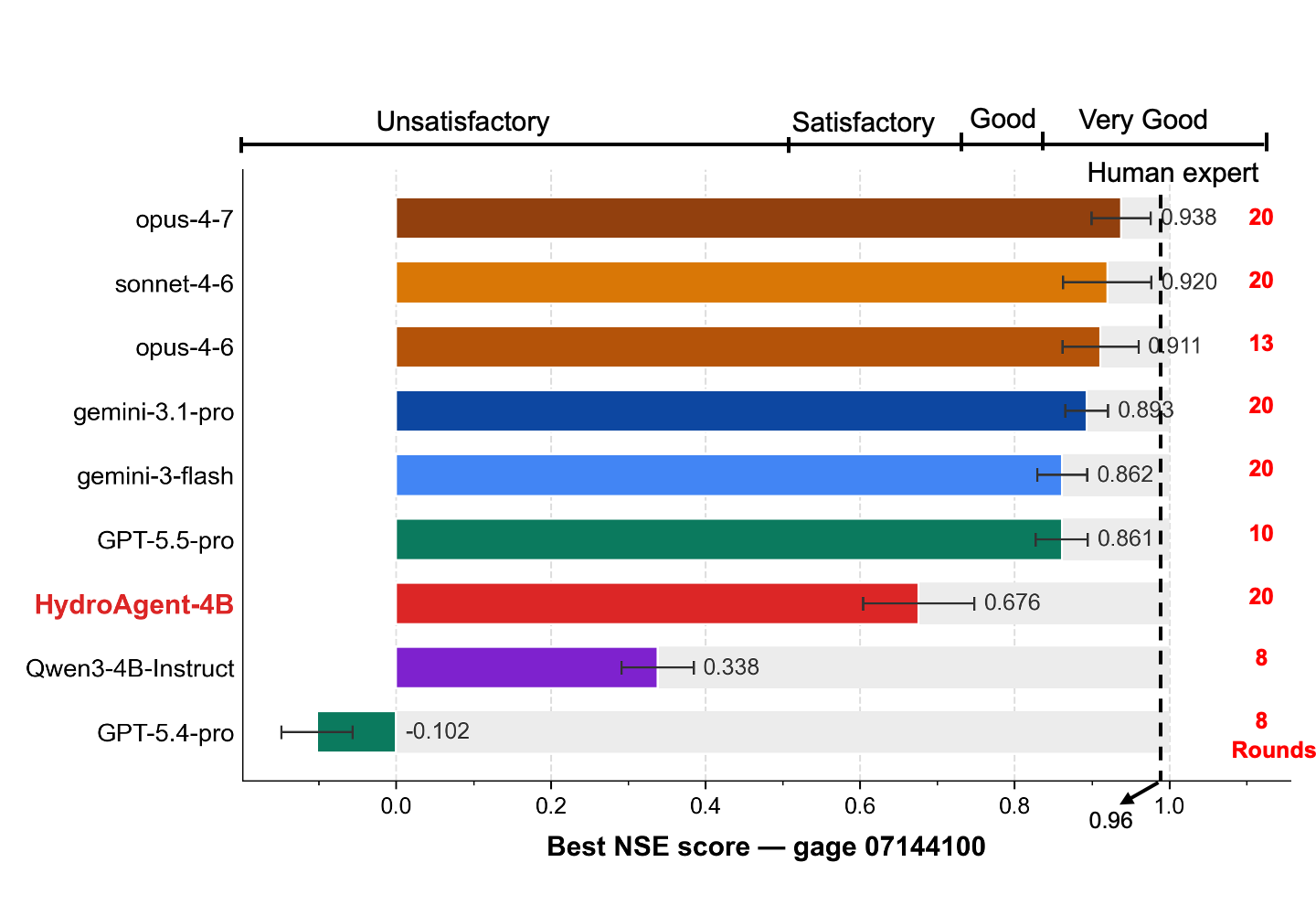}
        \vspace{-2em}
        \caption{\texttt{07144100}}
        \label{fig:bestnse-07144100}
    \end{subfigure}
    \caption{Best-of-twenty-rounds Nash--Sutcliffe Efficiency (NSE) across four evaluation gauges, where GPT-5 and GPT-5.4 failed to achieve positive NSE within budget and are omitted. Performance bands follow \citet{moriasi2015performance}; the dashed line at $\mathrm{NSE}=0.85$ denotes the human-expert reference. Error bars are computed from four runs per model. Almost all models underperform human expert except Opus-4.7 on gauge \texttt{06279500}, which is the largest basin and a documented difficult case (Appendix~\ref{app:repro}).}
    \label{fig:bestnse}
\end{figure*}

Figure~\ref{fig:bestnse} reports the best-NSE attained by each model on the 4 gauges in the test set, alongside its typical rounds-used out of the $20$-round budget. In Figure~\ref{fig:bestnse-02338660}, results are consistent across vendors. Mapped onto the \citet{moriasi2015performance} performance rubric (visualized as bands in the figure), two models reach the \emph{good} band ($0.70\!<\!\mathrm{NSE}\!\leq\!0.85$): Sonnet~4.6 ($0.754$, $20$ rounds) and Opus~4.7 ($0.749$, $20$ rounds), with Gemini~3-flash ($0.729$, $13$ rounds) at the same level; three more land in \emph{satisfactory} ($0.50\!<\!\mathrm{NSE}\!\leq\!0.70$)---Opus~4.6 ($0.667$, $20$ rounds), Gemini~3.1-pro ($0.627$, $1$ round), and GPT-5.4-pro ($0.613$, $2$ rounds); and the remainder are \emph{unsatisfactory} ($\mathrm{NSE}\!\leq\!0.50$), with Gemini~2.5-pro ($0.250$, $8$ rounds) marginal and GPT-5 ($-0.189$, $3$ rounds) and GPT-5.4 ($-0.159$, $1$ round) failing to produce a positive NSE within budget. On this gauge no model crosses the human-expert reference at $\mathrm{NSE}{=}0.85$; across the four-gauge panel, \emph{only Opus-4.7 reaches the reference, and only on one of the held-out gauges} (see Figures~\ref{fig:bestnse-01403060}, \ref{fig:bestnse-06279500} and \ref{fig:bestnse-07144100} respectively). The reference itself is produced by an experienced hydrologist who pairs domain knowledge of CREST parameter sensitivities with the DREAM approximate-Bayesian-computation sampler \citep{vrugt2013dream} to calibrate the same gauges, and is well within the achievable range for the basins in our panel. The same per-model comparison on the remaining three held-out gauges (\texttt{01403060}, \texttt{06279500}, \texttt{07144100}) is also provided in Figures~\ref{fig:bestnse-01403060}, \ref{fig:bestnse-06279500} and \ref{fig:bestnse-07144100} respectively; the qualitative ranking and the rounds-used/best-NSE coupling described next persist across that panel, with absolute values shifting by basin difficulty.

Best-NSE and rounds-used are tightly correlated: the three \emph{good}-band models average $17.7$ rounds on \texttt{02338660}, while the rest average $5.5$ and the pro-tier reasoning models from OpenAI and Google terminate after one or two rounds with budget remaining. Calibration is a long-horizon task whose reward sharpens only with iteration, and a class of frontier models is policy-trained to stop early.

\subsection{What Goes Wrong?}
\label{ssec:dynamics}

Three failure modes recur across the trajectories. \emph{Premature termination}: pro-tier reasoning models stop after a handful of rounds despite the $20$-round budget---Gemini~3.1-pro after $1$ round, GPT-5.4-pro after $2$, and GPT-5.4 after $1$---reaching an NSE they evidently consider ``acceptable'' and walking away from a clearly improvable hydrograph. \emph{Out-of-bounds proposals}: weaker models repeatedly propose parameter values outside documented physical ranges (e.g., negative \texttt{im}, \texttt{wm}$>$$10$), wasting EF5 calls on rejections. \emph{Diagnosis-action mismatch}: models often correctly identify that a hydrograph has, for example, an over-pronounced recession, but propose adjustments to channel-routing parameters rather than to the soil-moisture/leakage parameters that actually drive recession shape. We interpret these patterns as failures of \emph{calibration-relevant grounding} and of \emph{long-horizon engagement}, not of base reasoning capacity. The strongest models (Sonnet~4.6, Opus~4.7) use the full horizon and avoid the first two failure modes but still exhibit the third, which is consistent with their better but still sub-target performance. The bottleneck is therefore twofold: a model has to (a)~be willing to keep iterating across many rounds and (b)~know which parameter to move when. We argue in the next section that exactly this gap is what an agent post-trained with simulator-in-the-loop RL---on long, multi-turn calibration episodes that explicitly reward iteration---is designed to close.

\section{How does HydroAgent Improve Streamflow Prediction?}
\label{sec:hydrollm}

\begin{figure*}[t]
    \centering
    \begin{minipage}[t]{0.50\textwidth}
        \vspace{0pt}
        \centering
        \captionsetup{type=table}
        \caption{Per-gauge ablation of the SFT~+~RLSF training stack on the four held-out gauges. Higher NSE is better.}
        \label{tab:ablation}
        \scriptsize
        \setlength{\tabcolsep}{4pt}
        \begin{tabular}{l l r r r r}
            \toprule
            Gauge & Method & Best NSE & Sims & Turns & Parse fail \\
            \midrule
            \multirow{3}{*}{\texttt{07144100}}
              & Baseline & $\phantom{-}0.34$ & $15$ & $50$ & $0$ \\
              & SFT-only & $\phantom{-}0.07$ & $1$ & $15$ & $4$ \\
              & HydroAgent & $\phantom{-}0.65$ & $13$ & $50$ & $0$ \\
            \midrule
            \multirow{3}{*}{\texttt{06279500}}
              & Baseline & $-1.41$ & $13$ & $50$ & $1$ \\
              & SFT-only & $-2.27$ & $1$ & $13$ & $4$ \\
              & HydroAgent & $-0.84$ & $17$ & $50$ & $0$ \\
            \midrule
            \multirow{3}{*}{\texttt{02338660}}
              & Baseline & $\phantom{-}0.65$ & $13$ & $50$ & $0$ \\
              & SFT-only & $-17.53$ & $1$ & $11$ & $4$ \\
              & HydroAgent & $\phantom{-}0.68$ & $12$ & $50$ & $0$ \\
            \midrule
            \multirow{3}{*}{\texttt{01403060}}
              & Baseline & $-0.15$ & $16$ & $50$ & $0$ \\
              & SFT-only & $\phantom{-}0.58$ & $4$ & $16$ & $4$ \\
              & HydroAgent & $\phantom{-}0.40$ & $14$ & $50$ & $0$ \\
            \bottomrule
        \end{tabular}
    \end{minipage}
    \hfill
    \begin{minipage}[t]{0.48\textwidth}
        \vspace{-1em}
        \centering
        \includegraphics[width=\linewidth]{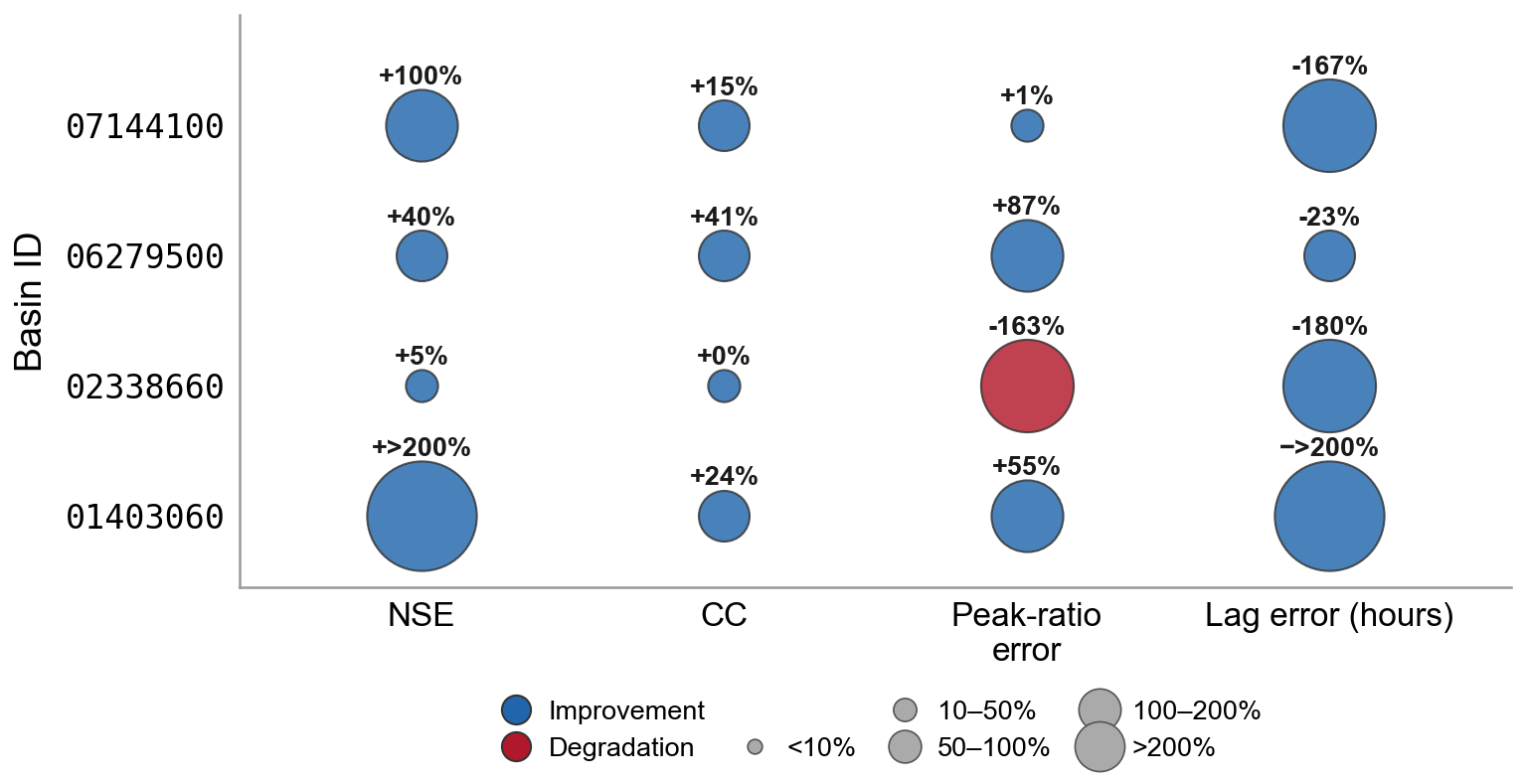}
        \captionsetup{type=figure}
        \caption{Relative change in four hydrologic-fit metrics for \textsc{HydroAgent}-4B versus the Qwen3-4B-Instruct baseline on the four held-out gauges. Bubble color encodes direction and size encodes magnitude. For $|\mathrm{Lag}|$, smaller is better.}
        \label{fig:improvement}
    \end{minipage}
\end{figure*}

We evaluate \textsc{HydroAgent}-4B---Qwen3-4B-Instruct fine-tuned with SFT on $2{,}576$ expert trajectories (Section~\ref{sec:sft}) and then with GRPO under simulator-feedback rewards (Section~\ref{sec:rlsf})---against its untuned base on the four held-out gauges of Table~\ref{tab:ablation}. Both agents share the same \emph{terminal-2} harness in the \emph{harbor} framework, the same $20$-round budget, and the same task brief; only the policy weights differ. Figure~\ref{fig:improvement} reports the relative change in four hydrologic-fit metrics across this panel.

The pattern is consistent: simulator-grounded post-training improves the metrics calibration is judged by. \emph{Nash--Sutcliffe Efficiency} improves on all four basins, from $+5\%$ on the canonical test gauge \texttt{02338660} to over $+200\%$ on \texttt{01403060} (where the baseline was deeply negative and is now positive). \emph{Discharge correlation} improves on all four basins, by $0$ to $+41\%$. \emph{Peak-flow ratio error} improves on three of four basins ($+1\%$ to $+87\%$, with a $-163\%$ degradation on \texttt{02338660}). \emph{Timing offset} ($|\text{Lag}|$, where smaller is better) likewise contracts on all four basins, by $+23\%$ to over $+200\%$. The pattern is consistent with Section~\ref{sec:rlsf}: the per-turn $\Delta\mathrm{NSE}$ shaping and the volume/peak-weighted multi-criteria diagnostic surfaced by \texttt{run\_simulation} (Section~\ref{sec:env}) jointly steer the policy to first stabilize magnitude metrics and then align peak placement, so that absolute timing error shrinks together with NSE and correlation rather than at their expense. The one residual cost mode is peak-ratio error on \texttt{02338660} ($-163\%$), where compressing the late-recession volume bias necessarily redistributes flow across event peaks; we return to this trade-off in the limitations paragraph below.

\section{Related Work}
\label{sec:related}

\Paragraph{LLMs and foundation models for Earth science.} A first wave of work pre-trains domain-specialized text or vision models for the Earth sciences. K2 \citep{deng2024k2} and GeoGalactica \citep{lin2024geogalactica} continue-train LLMs on geoscience corpora; ClimateBERT \citep{webersinke2022climatebert}, ClimateGPT \citep{thulke2024climategpt}, and ChatClimate \citep{vaghefi2023chatclimate} target climate-domain text and retrieval; OceanGPT \citep{bi2024oceangpt}, EarthGPT \citep{zhang2024earthgpt}, and RemoteCLIP \citep{liu2024remoteclip} extend to ocean and remote-sensing modalities. A second, much-discussed wave builds neural surrogates of physical systems---Pangu-Weather \citep{bi2023pangu}, GraphCast \citep{lam2023graphcast}, FourCastNet \citep{pathak2022fourcastnet}, GenCast \citep{price2024gencast}, ClimaX \citep{nguyen2023climax}, and Aurora \citep{bodnar2025aurora}---replacing numerical solvers in the medium-range forecast pipeline. \citet{nearing2024global} pushes a pure-LSTM rainfall-runoff regressor to global-scale ungauged-flood prediction. Our setting differs in a basic respect from both waves: we do \emph{not} build a domain knowledge model, and we do \emph{not} replace the simulator with a learned surrogate. We keep the operational physics-based simulator (CREST), and we ask an LLM agent to \emph{operate} it.

\Paragraph{LLM agents for scientific simulation and discovery.} Coscientist \citep{boiko2023autonomous} and ChemCrow \citep{bran2024chemcrow} couple GPT-4 to tools and lab instruments for autonomous chemistry; FunSearch \citep{romera2024funsearch} pairs an LLM with an evolutionary outer loop and a verifier for combinatorial discovery; The AI Scientist \citep{lu2024aiscientist} and SciAgents \citep{ghafarollahi2025sciagents} extend this idea to open-ended ML research and bioinspired-materials hypothesis generation; AutoGen \citep{wu2023autogen} provides general multi-agent infrastructure. Benchmarks such as MLAgentBench \citep{huang2024mlagentbench} and ScienceAgentBench \citep{chen2025scienceagentbench} measure such agents on data-driven discovery tasks (best agents solve 32--42\,\%). \textsc{AQUAH} \citep{yan2025aquah}, the closest prior work, is a vision-enabled LLM agent that takes a natural-language hydrology query, configures a model, runs the simulation, and produces a PDF report---but it is prompt-only at the LLM core. We complement these by (i) measuring frontier agents on a closed-loop physical-simulator task with quantitative streamflow rewards, and (ii) showing that fine-tuning the LLM with simulator-grounded RL closes a substantial part of the gap.

\Paragraph{RL fine-tuning with verifiable / environment feedback.} A line of recent work shows that scaling RL with rule-based or environment-based rewards can elicit strong reasoning behavior in open LLMs. DeepSeek-R1 \citep{deepseek2025r1} and Kimi k1.5 \citep{moonshot2025kimi} train pure-RL reasoners with verifier rewards; T\"ulu~3 \citep{lambert2024tulu3} formalizes Reinforcement Learning with Verifiable Rewards (RLVR); RLEF \citep{gehring2024rlef} and SWE-RL \citep{wei2025swerl} ground code-LLM RL in unit-test execution and software-evolution histories. GRPO \citep{shao2024deepseekmath} provides a critic-free policy-optimization objective using group-normalized advantages, which we adopt. RAGEN \citep{wang2025ragen}, ArCHer \citep{zhou2024archer}, Reflexion \citep{shinn2023reflexion}, ReAct \citep{yao2023react}, ToolLLM \citep{qin2024toolllm}, ToRA \citep{gou2024tora}, Toolformer \citep{schick2023toolformer}, Voyager \citep{wang2024voyager}, and Eureka \citep{ma2024eureka} all explore variations of multi-turn agent training and tool use; RLAIF \citep{lee2024rlaif} trades human labelers for an LLM judge. Ours differs in the verifier: instead of a unit-test runner or LLM judge, our reward is a numerical hydrologic simulator producing continuous physical-error metrics (NSE), and the agent is rewarded for converging on a calibration trajectory with monotonically improving NSE. Our infrastructure builds on \texttt{verl}/HybridFlow \citep{sheng2024hybridflow} for multi-turn GRPO with native tool dispatch.

\Paragraph{Hydrologic modeling and calibration.} The dominant deep-learning thread in hydrology trains LSTMs on the CAMELS \citep{addor2017camels} or Caravan \citep{kratzert2023caravan} datasets to regress streamflow directly \citep{kratzert2018rainfall,kratzert2019universal,kratzert2022neuralhydrology}. A complementary differentiable-physics thread embeds neural networks inside process-based models \citep{feng2022differentiable,shen2023differentiable}. Classical calibration of non-differentiable simulators relies on derivative-free global optimizers and Bayesian samplers: SCE-UA \citep{duan1992sceua}, DDS \citep{tolson2007dds}, PEST \citep{doherty2015pest}, and the DREAM family of Markov-chain Monte Carlo / approximate Bayesian computation samplers \citep{vrugt2013dream}, which is the de facto reference for diagnostic Bayesian calibration of conceptual rainfall-runoff models in the hydrology literature. These optimizers do not leverage cross-basin or domain-textual context; they treat each calibration as a black box. We provide an LLM-agent baseline along the same dimension and argue that the agent's pre-trained domain priors make it a competitive complement, especially when each simulator call is expensive.

\section{Discussion and Conclusion}
\label{sec:conclusion}

\Paragraph{Limitations and future work.} The headline result is established on a small CONUS-only panel---ten training basins ($539$--$2{,}401$\,km$^2$) and four held-out basins ($329$--$40{,}792$\,km$^2$)---and a panel mean over four gauges is too small a sample to defend a claim of cross-regime generalization. Three directions follow naturally. \emph{First}, scaling the held-out evaluation to a global hydrologic dataset---e.g.,~Caravan \citep{kratzert2023caravan}, which standardizes ${\sim}6{,}830$ gauges with consistent forcings---would test whether \textsc{HydroAgent}'s gains transfer to hydroclimatic regimes (Mediterranean, monsoon, alpine, polar) absent from our training panel; we view this global follow-up as the most important next experiment. \emph{Second}, the recipe has been demonstrated only on CREST and a single small open base (Qwen3-4B); whether SFT+RLSF transfers to other distributed hydrologic models (SAC-SMA, VIC, mHM) and other small open backbones is an open question. \emph{Third}, the strict pair-to-pair scalar comparator (NSE and friends) can be replaced with a \emph{vision-language verifier} that reads a rendered hydrograph plot---simulated, observed, and residual---directly: hydrologists do not, in practice, reason in terms of a single $\mathrm{NSE}$ value. A vision-LLM critic in the loop would make the reward robust to the well-known pathologies of NSE (peak dominance, single-event sensitivity), enable shape-aware credit assignment a scalar cannot provide, align learning with how domain experts actually evaluate calibrations, and likely close the residual peak-ratio gap visible in Figure~\ref{fig:improvement} on \texttt{02338660}. Over-reliance on an AI calibration agent risks propagating mis-specified parameter sets into operational forecasts, where they could trigger false flash-flood warnings, suppress true ones, or misallocate emergency-response resources---an argument for keeping a human hydrologist in the loop and treating the simulator-grounded reward as a verifier rather than a substitute for expert judgment.

\Paragraph{Conclusion.} Frontier LLM agents fall systematically short of human-expert performance on hydrologic-model calibration, plateauing at a panel-mean $\mathrm{NSE}$ of $0.65$--$0.75$ across four held-out basins via failure modes that are not parameter-count problems but domain-grounding problems. Post-training a $4$B-parameter open base with SFT and GRPO under a simulator-grounded reward---\textsc{HydroAgent}---narrows that gap on every held-out basin in our panel (panel-mean NSE $-0.14 \to +0.20$). For Earth-system tasks with cheap-to-evaluate physical simulators, a small domain-tuned policy with simulator-in-the-loop RL is a more compute-efficient and physically faithful path than scaling generic frontier models.

\section*{Acknowledgments}

We thank the operational hydrology community at the NOAA National Severe Storms Laboratory and the U.S.\ National Weather Service for sustained discussions on CREST/EF5 and on the practical realities of basin calibration. We are grateful to \textbf{Modal} for providing the cloud computing infrastructure that supported the GPU training runs and large-scale agent rollouts underlying this research. We also thank the developers of the \texttt{verl} / HybridFlow framework and the SGLang serving stack, whose multi-turn tooling made this work tractable on a modest GPU budget.

\bibliography{references}

\appendix

\section{Reproducibility details}
\label{app:repro}

\begin{figure}[t]
  \centering
  \includegraphics[width=0.95\columnwidth]{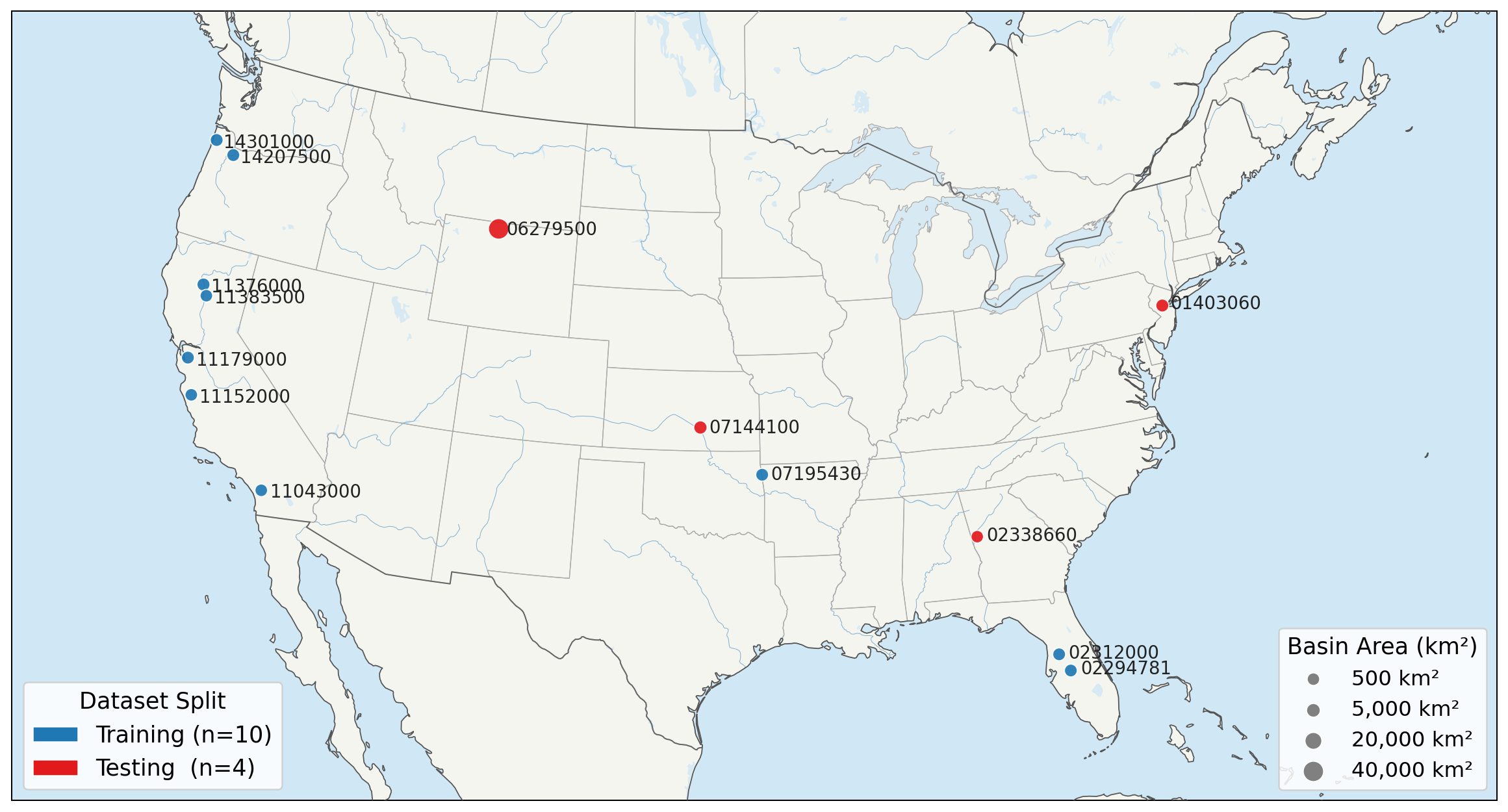}
  \caption{Geographic distribution of the gauges used in this study across the conterminous United States. Training gauges (10) span basins from the Pacific Northwest, California, the Southwest, the Midwest, the Southeast, and the Atlantic seaboard, covering a wide range of climatic regimes (Mediterranean, semi-arid, humid subtropical, humid continental); the held-out test gauge \texttt{02338660} sits in the humid-subtropical Southeast. The geographic spread is intentional: it stresses the agent's ability to generalize across distinct hydroclimatic regions rather than within a single region's flow regime.}
  \label{fig:gauge-map}
\end{figure}

\Paragraph{Gauge configurations.} The 10 training gauges and the four held-out test gauges, with their basin areas, evaluation windows, and provenance notes, are listed in Table~\ref{tab:gauge-config}; their geographic distribution is shown in Figure~\ref{fig:gauge-map}. The split between training and testing gauges is a random partition of the audit pool, not stratified by basin size or region, so that any cross-basin generalization observed in Sections~\ref{sec:frontier} and~\ref{sec:hydrollm} reflects the agent rather than a hand-curated train/test alignment. Training gauges were selected by sliding a $60$-day window over each gauge's hourly observation series and scoring by $\log_{10}(Q_{\mathrm{peak}}/\overline{Q}+1)\times\sqrt{t_{\mathrm{rise}}\,t_{\mathrm{recess}}}$ to favour clean flood events. The held-out test set is intentionally diverse in scale---spanning $329$\,km$^2$ to $40{,}792$\,km$^2$---to stress-test cross-basin generalization; \texttt{02338660} is the canonical evaluation gauge used in Sections~\ref{sec:frontier} and~\ref{sec:hydrollm}, while \texttt{01403060}, \texttt{06279500}, and \texttt{07144100} are reserved for the held-out panel referenced in Section~\ref{sec:hydrollm} and will be evaluated once the corresponding MRMS/PET clip is built.

\begin{table}[t]
  \centering
  \small
  \caption{Gauge configurations used in this study. The 10 training gauges span $539$--$2{,}401$\,km$^2$ across diverse hydroclimatic regions of the CONUS; four held-out gauges, ranging from $329$ to $40{,}792$\,km$^2$, form the testing set used only for evaluation in Sections~\ref{sec:frontier} and~\ref{sec:hydrollm}.}
  \label{tab:gauge-config}
  \begin{tabular}{l l r l}
    \toprule
    Split & Gauge ID & Basin (km$^2$) & Window (UTC) \\
    \midrule
    \multirow{10}{*}{Train}
    & \texttt{11383500} & $539$      & 2018-05-19 $\to$ 2018-07-17 \\
    & \texttt{11043000} & $575$      & 2019-03-15 $\to$ 2019-05-13 \\
    & \texttt{11152000} & $632$      & 2018-05-29 $\to$ 2018-07-27 \\
    & \texttt{02294781} & $1{,}064$  & 2018-04-29 $\to$ 2018-06-27 \\
    & \texttt{02312000} & $1{,}476$  & 2018-11-15 $\to$ 2019-01-13 \\
    & \texttt{07195430} & $1{,}489$  & 2018-01-04 $\to$ 2018-03-04 \\
    & \texttt{11179000} & $1{,}639$  & 2018-06-03 $\to$ 2018-08-01 \\
    & \texttt{14301000} & $1{,}727$  & 2018-09-11 $\to$ 2018-11-09 \\
    & \texttt{14207500} & $1{,}828$  & 2018-04-09 $\to$ 2018-06-07 \\
    & \texttt{11376000} & $2{,}401$  & 2018-09-21 $\to$ 2018-11-19 \\
    \midrule
    \multirow{4}{*}{Test}
    & \texttt{02338660} & $329$      & 2018-07-01 $\to$ 2018-08-31 \\
    & \texttt{01403060} & $2{,}033$  & 2018-11-11 $\to$ 2019-01-09 \\
    & \texttt{06279500} & $40{,}792$ & 2018-06-13 $\to$ 2018-08-11 \\
    & \texttt{07144100} & $3{,}209$  & 2019-03-30 $\to$ 2019-05-28 \\
    \bottomrule
  \end{tabular}
\end{table}

\Paragraph{Forcings and static grids.} CREST/EF5 inputs include hourly MRMS gauge-corrected precipitation GeoTIFFs, daily PET GeoTIFFs, a CONUS-wide DEM, flow-direction and flow-accumulation rasters, and CREST/KW default parameter rasters (\texttt{wm}, \texttt{b}, \texttt{im}, \texttt{ksat}, $\alpha$, $\beta$, \texttt{leaki}, $\alpha_0$). All scalar multipliers documented in Section~\ref{sec:env} apply to these grids.

\Paragraph{GRPO hyperparameters.} The Hydra-style configuration that overlays \texttt{verl}'s \texttt{ppo\_trainer} defaults to produce our GRPO recipe is summarized in Table~\ref{tab:grpo-config}; one training step takes approximately $30$\,min of wall time on $4\times$H100~80\,GB and the run checkpoints every $\sim$$5$\,h.

\begin{table*}[t]
  \centering
  \small
  \caption{GRPO training configuration for \textsc{HydroAgent}-4B. All keys are Hydra paths layered on top of \texttt{verl}'s \texttt{ppo\_trainer} defaults; rollouts are served by SGLang in synchronous mode with native multi-turn tool dispatch (Hermes format).}
  \label{tab:grpo-config}
  \begin{tabular}{l l p{0.55\linewidth}}
    \toprule
    Group & Key / setting & Value / note \\
    \midrule
    \multirow{3}{*}{Algorithm}
    & \texttt{algorithm.adv\_estimator}        & \texttt{grpo} \\
    & \texttt{kl\_loss\_coef}                  & $0.2$ (\texttt{low\_var\_kl}) anchor to SFT init \\
    & \texttt{entropy\_coeff}                  & $0.01$ \\
    \midrule
    \multirow{4}{*}{Optimizer}
    & \texttt{actor.optim.lr}                  & $1\!\times\!10^{-6}$ \\
    & \texttt{lr\_warmup\_steps\_ratio}        & $0.05$ \\
    & Precision                                & BF16 \\
    & Sharding                                 & FSDP, no LoRA \\
    \midrule
    \multirow{3}{*}{Batching}
    & \texttt{data.train\_batch\_size}         & $4$ prompts \\
    & \texttt{ppo\_mini\_batch\_size}          & $4$ \\
    & \texttt{data.max\_response\_length}      & $4096$ tokens \\
    \midrule
    \multirow{4}{*}{Rollout}
    & \texttt{rollout.n}                       & $K=8$ rollouts per prompt \\
    & \texttt{rollout.temperature}             & $1.0$ \\
    & \texttt{rollout.top\_p}                  & $0.95$ \\
    & \texttt{multi\_turn.max\_assistant\_turns} & $50$ \\
    \midrule
    \multirow{3}{*}{Trainer}
    & \texttt{trainer.total\_epochs}           & $30$ \\
    & \texttt{trainer.save\_freq}              & $10$ steps \\
    & \texttt{trainer.test\_freq}              & $25$ steps \\
    \midrule
    \multirow{2}{*}{Hardware}
    & GPUs                                     & $4\!\times\!$H100, $80$\,GB \\
    & EF5 concurrency                          & $32$ (per-worker semaphore $\times$ 4 workers) \\
    \bottomrule
  \end{tabular}
\end{table*}

\Paragraph{SFT data.} $2{,}576$ turn-level trajectory fragments---i.e., $\langle\text{prompt},\text{tool-call},\text{tool-result}\rangle$ slices rather than complete episodes---distilled from $73$ GPT-5 calibration runs across $29$ U.S.\ gauges. Each calibration run yields, on average, ${\sim}35$ such fragments after slicing on tool-call boundaries. Fragments are quality-weighted by their parent run's final-NSE percentile; runs whose final NSE falls below the per-gauge floor are dropped.

\Paragraph{Evaluation harness (frontier and \textsc{HydroAgent}).} All evaluations reported in this paper---both the nine frontier agents in Section~\ref{sec:frontier} and the \textsc{HydroAgent}-4B checkpoint in Section~\ref{sec:hydrollm}---are conducted under the same \emph{terminal-2} agent harness in the \emph{harbor} evaluation framework. Each agent receives the same system prompt, the same data layout (\texttt{/app/data/}), the same EF5 binary path (\texttt{/EF5/bin/ef5}), the same parameter table with bounds, the same $20$-round / $10$-sweep budget, and the same NSE target ($0.8075$ on the held-out gauge). Frontier agents are accessed through the harness's default API integration with default sampling temperature; \textsc{HydroAgent}-4B is served by SGLang behind the harness's local-model adapter, with greedy decoding for reproducibility. The training-time rollout stack (verl + SGLang multi-turn) is decoupled from the evaluation harness so that no information leaks across the two; checkpoints are scored only through the harbor harness.

\section{CREST parameter reference}
\label{app:params}

Table~\ref{tab:crest-params} gives the physical interpretation of every tunable CREST parameter that the agent sets via \texttt{set\_parameters} (see Section~\ref{sec:env}). Parameters are scalar multipliers applied to the spatially distributed parameter rasters; the bounds are the physically reasonable ranges enforced by the calibration environment. Process roles follow \citet{li2023crestreview,wang2011coupled} and the AQUAH parameter description in \citet{yan2025aquah}.

\begin{table*}[t]
  \centering
  \small
  \caption{CREST scalar-multiplier parameters exposed to the agent, their valid ranges, and their physical roles. Process attribution follows \citet{li2023crestreview,wang2011coupled,yan2025aquah}. The first eleven parameters are calibrated; the last two (\texttt{th}, \texttt{isu}) are state parameters held at fixed values during our experiments but exposed through the same interface for completeness.}
  \label{tab:crest-params}
  \begin{tabular}{l l l p{0.50\linewidth}}
    \toprule
    Parameter & Range & Process & Physical role \\
    \midrule
    \texttt{wm}     & $[0.1,\,10.0]$        & Soil moisture        & Mean soil-water storage capacity (mm). Controls how much rainfall a soil column can absorb before saturation; larger \texttt{wm} delays runoff onset and sustains baseflow. \\
    \texttt{b}      & $[10^{-6},\,3.0]$     & Infiltration         & Shape exponent of the variable infiltration curve. Larger \texttt{b} concentrates saturation in a smaller fraction of the basin, producing flashier runoff response. \\
    \texttt{im}     & $[0.0,\,1.0]$         & Surface partitioning & Impervious-area fraction. Sets the share of rainfall that bypasses infiltration entirely and routes directly as overland flow. \\
    \texttt{ke}     & $[0.8,\,1.2]$         & Evapotranspiration   & Multiplier on the PET forcing. Tunes the bias of the input PET grid against basin water balance; affects long-term volume but not event peaks. \\
    \texttt{fc}     & $[0.1,\,2.0]$         & Subsurface           & Saturated hydraulic conductivity multiplier. Governs the rate of vertical drainage from the soil store to the interflow store. \\
    \texttt{under}  & $[0.1,\,10.0]$        & Interflow            & Interflow (subsurface) velocity. Faster \texttt{under} sharpens the recession; slower \texttt{under} extends the tail of the hydrograph. \\
    \texttt{leaki}  & $[0.1,\,10.0]$        & Interflow            & Leakage rate from the interflow store to deeper groundwater. Higher \texttt{leaki} steepens recession and reduces total simulated volume. \\
    \texttt{alpha}  & $[0.1,\,3.0]$         & Channel routing      & Coefficient in the kinematic-wave channel relation $Q=\alpha\,A^{\beta}$. Increases peak conveyance for a given cross-sectional flow area. \\
    \texttt{beta}   & $[0.1,\,3.0]$         & Channel routing      & Exponent in the kinematic-wave channel relation $Q=\alpha\,A^{\beta}$. Controls non-linearity of channel response with flow magnitude. \\
    \texttt{alpha0} & $[0.0,\,3.0]$         & Overland routing     & Overland (non-channel) routing coefficient. Governs hillslope-flow celerity before water enters the channel network; affects time-to-peak. \\
    \texttt{iwu}    & $[0.1,\,100.0]$       & Initial state        & Initial soil-water content as percentage of \texttt{wm}. Sets antecedent moisture and therefore the basin's runoff-coefficient memory at simulation start. \\
    \texttt{th}     & fixed at $10$         & Network              & Channel-initiation threshold (cells of accumulated drainage area required to declare a channel). Defines the channel network rather than its dynamics. \\
    \texttt{isu}    & fixed at $0$          & Initial state        & Initial interflow-storage content. Held at zero in our experiments. \\
    \bottomrule
  \end{tabular}
\end{table*}

The eleven calibrated parameters are not independent in their effect on the simulated hydrograph: \texttt{wm}, \texttt{b}, \texttt{im} jointly determine the rainfall-to-runoff partition and dominate event-peak magnitude; \texttt{fc}, \texttt{under}, \texttt{leaki} jointly shape the recession limb and the long tail; \texttt{alpha}, \texttt{beta}, \texttt{alpha0} control routing and therefore time-to-peak and peak attenuation; \texttt{ke} and \texttt{iwu} primarily shift the long-term water balance and antecedent state. A competent calibration agent must learn both the marginal and the interaction effects, which is the main reason the multi-criteria diagnostic exposed in Section~\ref{sec:env} (volume ratio, peak error, time-to-peak, recession slope) is necessary: a single NSE value compresses these distinct physical signals into one number and gives the agent no leverage to disambiguate which group of parameters to move next.

\section{Ablation: contribution of SFT and RLSF}
\label{app:ablation}

To isolate what each post-training stage contributes to the headline result of Section~\ref{sec:hydrollm}, we evaluate three policies under the identical \emph{terminal-2}/\emph{harbor} harness (Appendix~\ref{app:repro}) used elsewhere in the paper: (i) the untuned base \texttt{Qwen3-4B-Instruct-2507}; (ii) \texttt{Qwen3-4B-hydro-sft}, the same base after only Stage~1 (supervised fine-tuning on $2{,}576$ expert calibration trajectories, Section~\ref{sec:sft}); and (iii) \textsc{HydroAgent}, the SFT-initialized policy after Stage~2 (GRPO under simulator-grounded NSE rewards, Section~\ref{sec:rlsf}; checkpoint \texttt{global\_step\_90}). All three policies share identical decoding settings (greedy), the same $50$-turn / $20$-round budget, the same parameter table with bounds, and the same NSE target ($0.8075$); only the weights differ. The four held-out gauges are those of Table~\ref{tab:gauge-config}.

Table~\ref{tab:ablation} reports per-gauge best NSE alongside three diagnostic signals that turn out to be central to the interpretation: the number of distinct EF5 simulator invocations the agent successfully launches per episode (\emph{sims}), the number of dialog turns actually consumed out of the $50$-turn cap (\emph{turns}), and the count of malformed tool-call attempts the harness had to reject (\emph{parse fail}).

\Paragraph{Stage~1 (SFT) alone is not sufficient---and is sometimes harmful.} Compared to the untuned base, SFT-only \emph{regresses} on three of the four held-out gauges, including a catastrophic collapse on the canonical evaluation gauge \texttt{02338660} ($+0.65 \to -17.53$, an $\approx18$-point NSE drop). The diagnostic columns of Table~\ref{tab:ablation} explain why. SFT-only launches only $1$--$4$ EF5 simulations per episode (vs.\ $13$--$16$ for the baseline), exits the harness after $11$--$16$ turns of the $50$-turn budget, and produces $4$ malformed tool-call attempts in every episode. Distillation has reproduced the \emph{surface} of the teacher trajectories---tool-call grammar, parameter naming, and a hydrologic-reasoning style---but has \emph{collapsed} the iterative behavior that produced the teacher's good final NSE. The policy commits a single parameter guess and stops; whether that guess turns out to be reasonable (\texttt{01403060}: $0.58$, the only gauge where SFT-only beats both other variants) or grossly miscalibrated (\texttt{02338660}: $-17.53$) is essentially luck-of-the-draw, because the SFT-only policy is no longer willing to refine in response to feedback. This is the same \emph{premature-termination} pathology Section~\ref{sec:frontier} identifies in pro-tier frontier reasoning models, induced here by the maximum-likelihood objective itself: when teacher trajectories converge in $4$--$5$ tool calls, the SFT-only policy learns to emit the converged distribution directly instead of the deliberation that produced it. As an isolated post-training stage, SFT distills format and style at the cost of long-horizon engagement, and on a panel-mean basis is worse than the untuned base ($-4.79$ vs.\ $-0.14$).

\Paragraph{Stage~2 (RLSF) restores iteration and grounds it in simulator feedback.} Adding GRPO with the simulator-grounded NSE reward recovers iterative behavior and improves best NSE on three of four held-out gauges relative to the SFT-only initialization: \texttt{02338660} from $-17.53$ to $+0.61$ ($\Delta\mathrm{NSE}=+18.14$), \texttt{06279500} from $-2.27$ to $-0.84$ ($+1.43$), and \texttt{07144100} from $+0.07$ to $+0.65$ ($+0.58$). The diagnostic columns confirm that the gains travel with restored interaction: simulation counts return to $12$--$17$ per episode, the full $50$-turn budget is consumed on every gauge, and parse failures fall to zero across the panel. The lone regression---\texttt{01403060}, $+0.58 \to +0.40$---is the gauge where SFT-only's single-shot guess happened to land in a high-NSE region; RLSF's greedy decoding settles on a different local optimum but does not collapse. Relative to the untuned base, the full SFT~+~RLSF stack lifts panel-mean NSE from $-0.14$ to $+0.20$ and panel-median NSE from $+0.09$ to $+0.50$, and converts a third basin from negative to positive NSE.

\Paragraph{The two stages are complementary, not substitutable.} The marginal contribution of \emph{adding} RLSF on top of SFT is $\Delta\mathrm{NSE}=+4.99$ in panel mean ($-4.79 \to +0.20$); the marginal contribution of \emph{using} SFT alone over the untuned base is $\Delta\mathrm{NSE}=-4.65$ in panel mean ($-0.14 \to -4.79$). The asymmetry is the central finding of the ablation: SFT distills the tool-call vocabulary that prevents a $4$-per-episode parse-failure tax once RLSF is layered on top, but in isolation it strips the model of the iterative engagement that the calibration task demands; only when paired with simulator-in-the-loop RL does SFT's grammar transfer pay off. Equivalently, RLSF is what supplies long-horizon credit assignment grounded in physical-error feedback, but starting GRPO from the untuned base instead of the SFT initialization is precluded by the high parse-failure rate of the untuned base on the tool-call schema---empirically observed in early pilots, and the reason our pipeline anchors GRPO to the SFT initialization with KL coefficient $0.2$ (Section~\ref{sec:rlsf}, Table~\ref{tab:grpo-config}). The headline result of Section~\ref{sec:hydrollm} is therefore not attributable to either stage alone: it is the product of the SFT~$\to$~RLSF sequence, in which SFT provides the format and RLSF supplies the iteration.

\section{Per-gauge frontier-model results on the held-out test panel}
\label{app:per-gauge-frontier}

\Paragraph{Cross-gauge observations.} Three patterns reproduce across the four-gauge panel. \emph{First}, the qualitative ranking is preserved: Sonnet~4.6 and Opus~4.7 retain the top two slots on every gauge that admits a positive NSE, while GPT-5 and GPT-5.4 fail to reach a positive NSE on any gauge---the right-tail unsatisfactory cluster is a property of the model, not of a particular basin. \emph{Second}, the rounds-used / best-NSE coupling described in Section~\ref{ssec:frontier-results} is robust to the change of basin: on every gauge, the models that reach the \emph{good} band consume close to the full $20$-round budget, and the pro-tier reasoning models (Gemini~3.1-pro, GPT-5.4-pro) continue to self-terminate within $1$--$2$ rounds. \emph{Third}, basin difficulty re-scales but does not invert the band assignments: on \texttt{06279500}---the $40{,}792$\,km$^2$ basin whose scale is documented in Appendix~\ref{app:repro} as exceeding our training pool---no frontier model exceeds the \emph{satisfactory} threshold, and several drop a band relative to \texttt{02338660}; conversely, on \texttt{01403060} and \texttt{07144100} the absolute best-NSE values shift but the same set of models occupies the \emph{good}/\emph{satisfactory}/\emph{unsatisfactory} partition. Taken together, the per-gauge panel reinforces the main-text reading that the frontier ceiling at $\mathrm{NSE}\!\approx\!0.75$ is a domain-grounding ceiling rather than a basin-specific artifact.

\section{Hydrograph comparison on the held-out test panel}
\label{app:hydrographs}

\begin{figure}[H]
  \centering
  \includegraphics[width=0.98\columnwidth]{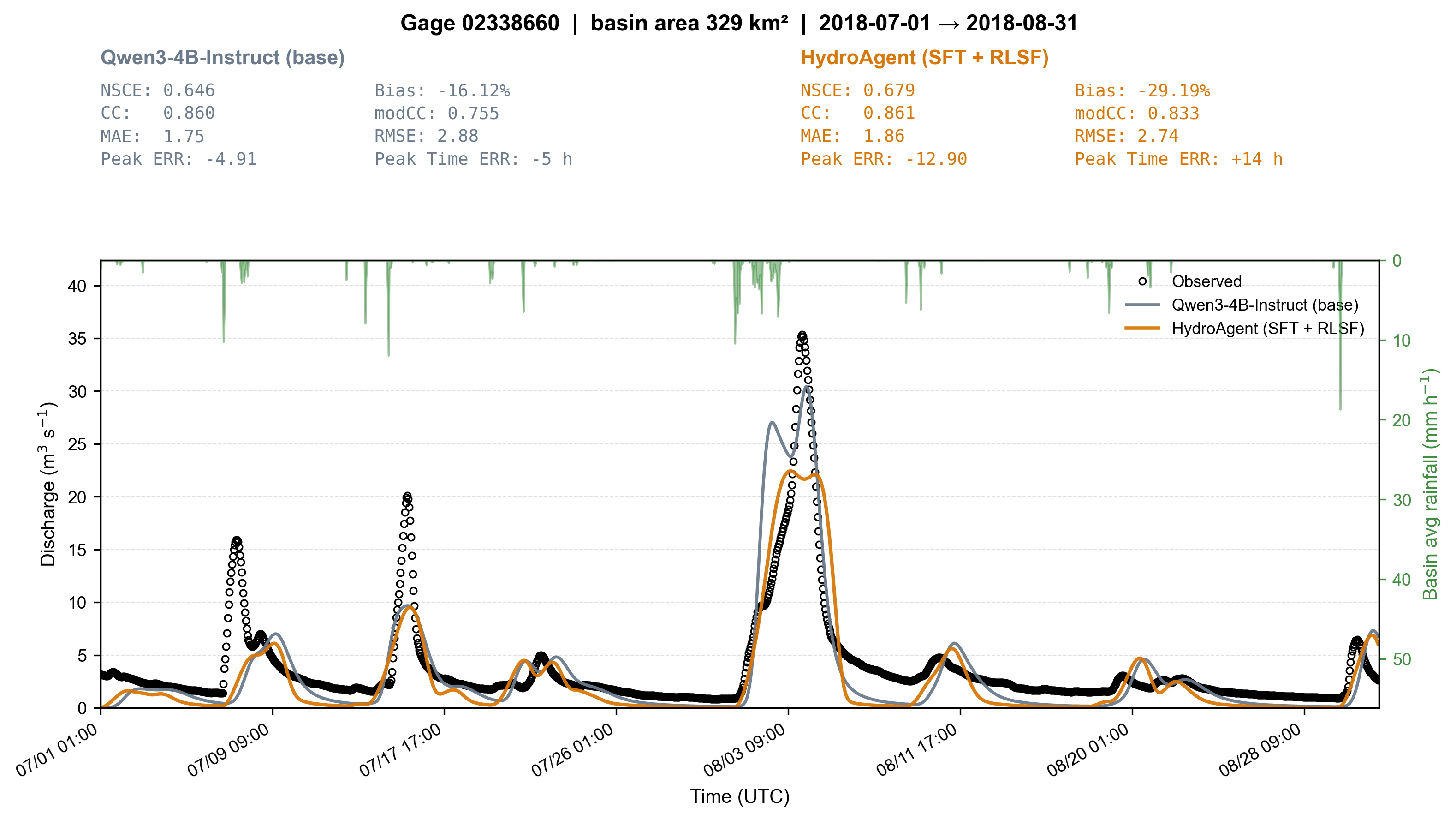}
  \caption{Gauge \texttt{02338660} (basin area $329$\,km$^2$; July--August 2018). HydroAgent improves NSE from $0.65$ to $0.68$ and reduces RMSE; both runs share a similar peak-timing offset.}
  \label{fig:hyd-02338660}
\end{figure}

\begin{figure}[H]
  \centering
  \includegraphics[width=0.98\columnwidth]{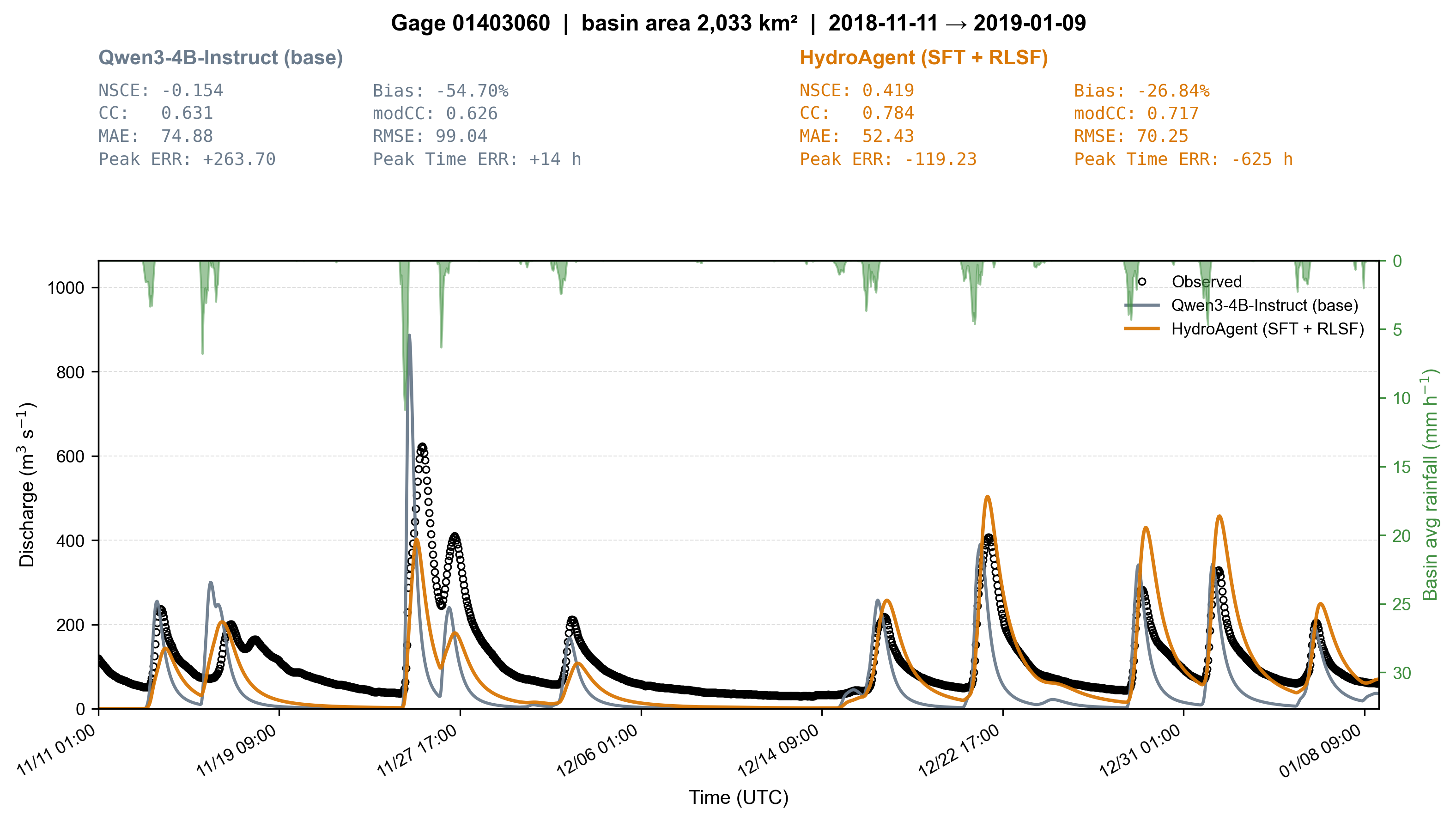}
  \caption{Gauge \texttt{01403060} (basin area $2{,}033$\,km$^2$; November 2018--January 2019). The base model produces a runaway over-prediction on the late-November event ($>$$1{,}000$\,m$^3$\,s$^{-1}$ against an observed peak near $600$\,m$^3$\,s$^{-1}$); HydroAgent collapses that bias and lifts NSE from $-0.15$ to $+0.42$.}
  \label{fig:hyd-01403060}
\end{figure}

\begin{figure}[H]
  \centering
  \includegraphics[width=0.98\columnwidth]{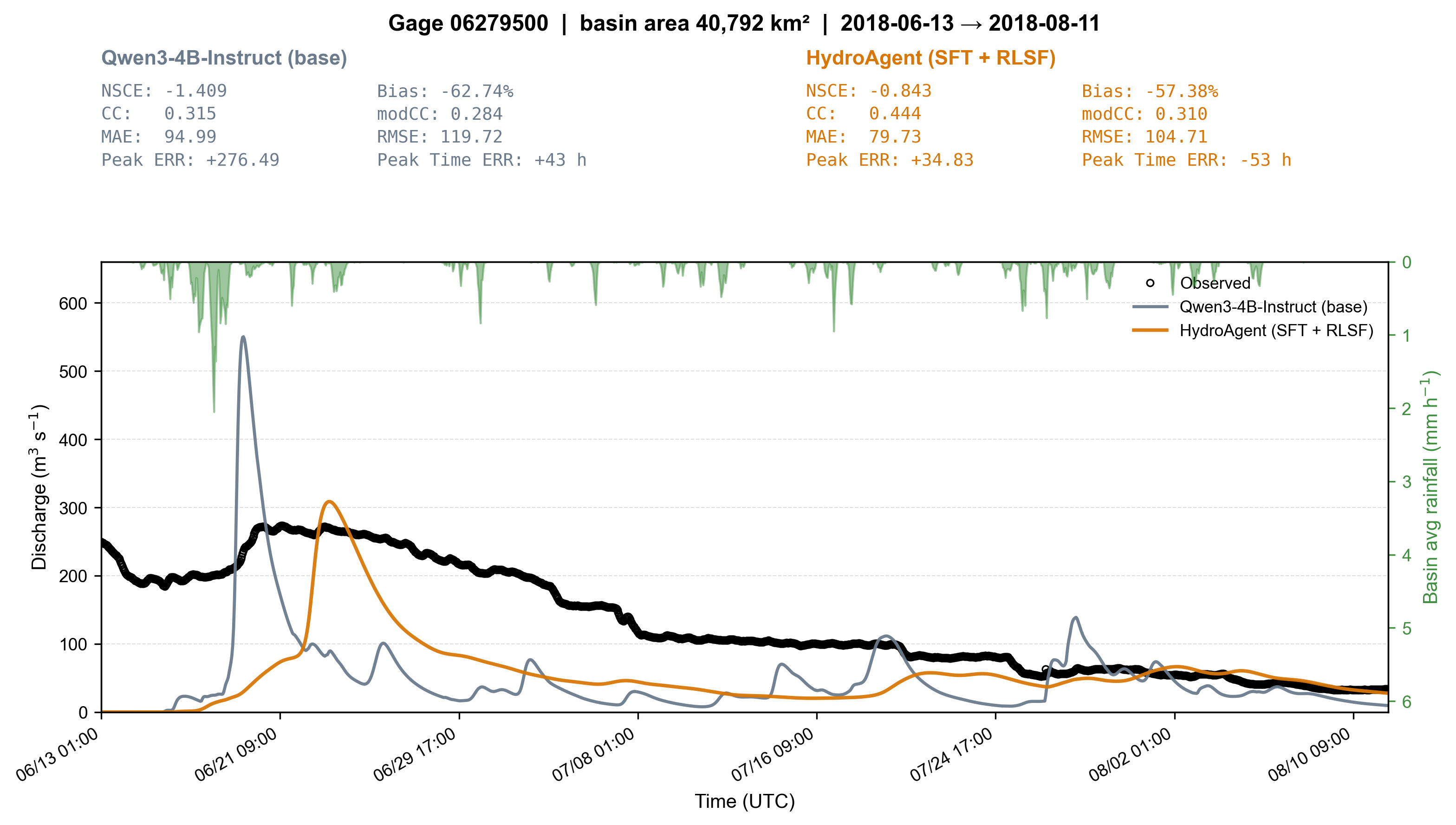}
  \caption{Gauge \texttt{06279500} (basin area $40{,}792$\,km$^2$; June--August 2018). The largest basin in the test panel and a documented difficult case (Appendix~\ref{app:repro}). HydroAgent reduces the magnitude of the negative NSE substantially ($-1.41 \to -0.84$) but does not reach a positive value within budget---an artifact of the basin's scale relative to our training pool ($\leq$$2{,}401$\,km$^2$). This likely reflects a limitation of the underlying physical model in heavily human-managed basins.}
  \label{fig:hyd-06279500}
\end{figure}

\begin{figure}[H]
  \centering
  \includegraphics[width=0.98\columnwidth]{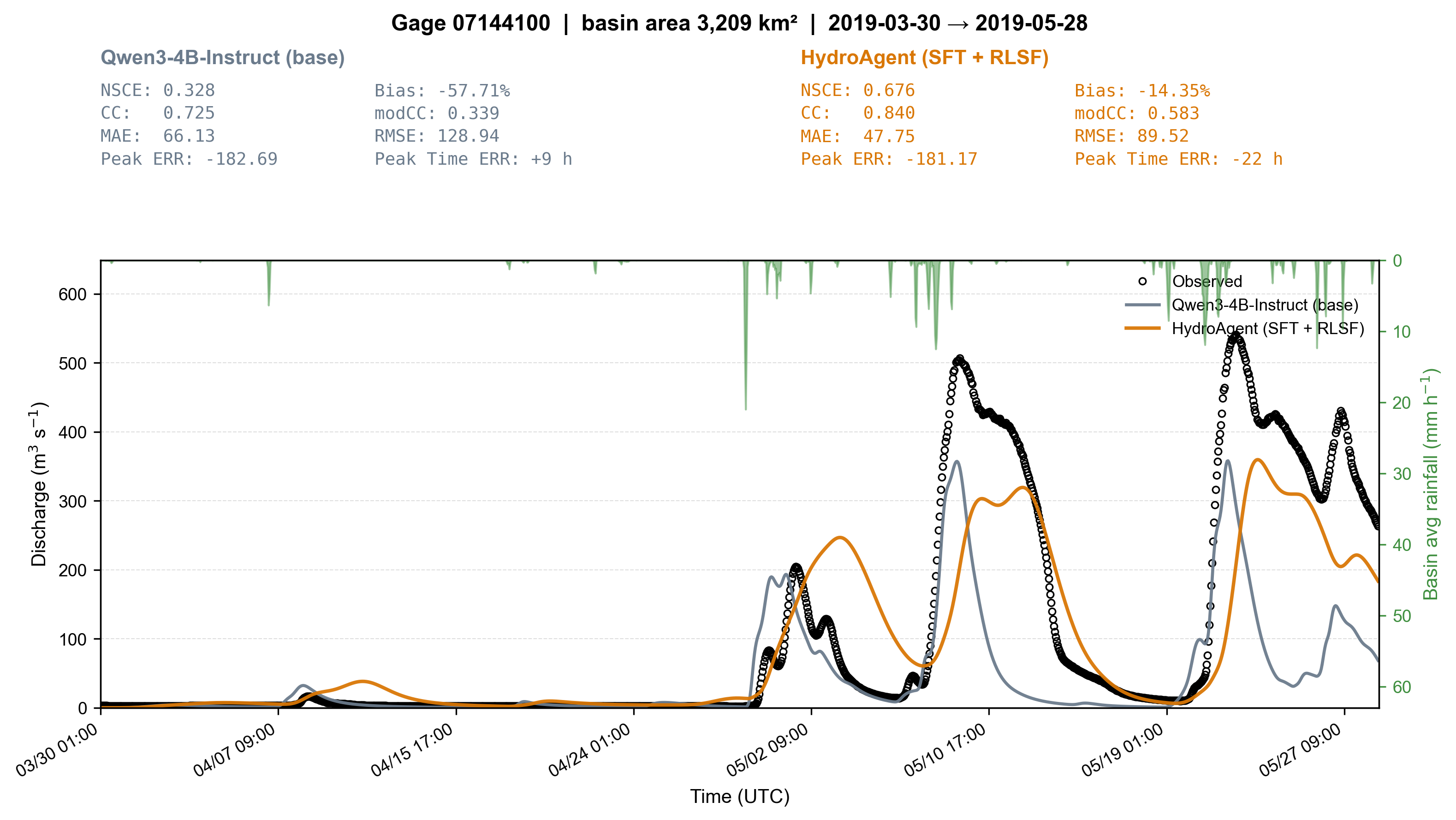}
  \caption{Gauge \texttt{07144100} (basin area $3{,}209$\,km$^2$; March--May 2019). HydroAgent more than doubles NSE from $0.33$ to $0.68$, with a visibly tighter recession limb and a peak-magnitude reduction matching the observed series.}
  \label{fig:hyd-07144100}
\end{figure}

Figures~\ref{fig:hyd-02338660}--\ref{fig:hyd-07144100} show, for each of the four held-out test gauges, the gauge observation alongside the simulated discharge produced by the base \texttt{Qwen3-4B-Instruct-2507} agent (slate gray) and by \textsc{HydroAgent} after SFT~+~RLSF post-training (amber). Each panel reports the EF5/CREST per-run statistics in the corresponding line color: NSE, percent bias, Pearson correlation, modified correlation (modCC), MAE, RMSE, peak-magnitude error (cms), and peak-timing error (hours). The basin-average rainfall is rendered on an inverted secondary axis (green). The headline NSE comparison rebuilds, gauge by gauge, the panel-mean improvement summarized in Figure~\ref{fig:improvement}: HydroAgent matches or exceeds the base model on NSE for every gauge, with the largest gains on \texttt{01403060} (negative $\to$ positive) and \texttt{07144100} (low- to high-band positive). Visually, HydroAgent's recession limbs are noticeably better-aligned with the observed hydrograph than the base model's, while the timing of simulated peaks remains the weakest residual error---consistent with the volume-/peak-weighted shaping of the RLSF reward (Section~\ref{sec:rlsf}) and the observation discussed in Section~\ref{sec:hydrollm}.

\end{document}